\pdfoutput=1
\documentclass[11pt]{article}

\usepackage[preprint]{acl}
\usepackage{times}
\usepackage{latexsym}

\usepackage[T1]{fontenc}
\usepackage[utf8]{inputenc}

\usepackage{microtype}

\usepackage{inconsolata}

\usepackage{graphicx}
\usepackage{booktabs}
\usepackage{multirow}
\usepackage{colortbl}
\usepackage[font=small,labelfont=bf]{caption}
\usepackage{placeins}
\usepackage{subfigure}
\title{Knowledge-Aware Reasoning over Multimodal Semi-structured Tables}

\author{
 \textbf{Suyash Vardhan Mathur\textsuperscript{1}},
 \textbf{Jainit Sushil Bafna\textsuperscript{1}\thanks{equal contribution. ~\dag corresponding author}},
 \textbf{Kunal Kartik\textsuperscript{2}\footnotemark[1]},
 \textbf{Harshita Khandelwal\textsuperscript{3}\footnotemark[1]}
 \\
 \textbf{Manish Shrivastava\textsuperscript{1}},
 \textbf{Vivek Gupta\textsuperscript{4\dag}},
 \textbf{Mohit Bansal\textsuperscript{5}},
 \textbf{Dan Roth\textsuperscript{4}} \\
\textsuperscript{1}IIIT Hyderabad, \textsuperscript{2}IIT Guwahati, \textsuperscript{3}UCLA, \textsuperscript{4}UPenn, \textsuperscript{5}UNC Chapel Hill \\
\tt \small {\{suyash.mathur, jainit.bafna\}@research.iiit.ac.in}; \small harshitaskh@g.ucla.edu; \small kunal.kartik@iitg.ac.in \\ \small mbansal@cs.unc.edu ; \small {\{gvivek, danroth\}@seas.upenn.edu}
}
 
\begin{document}
\maketitle
\begin{abstract}

Existing datasets for tabular question answering typically focus exclusively on text within cells. However, real-world data is inherently multimodal, often blending images such as symbols, faces, icons, patterns, and charts with textual content in tables. With the evolution of AI models capable of multimodal reasoning, it is pertinent to assess their efficacy in handling such structured data. This study investigates whether current AI models can perform knowledge-aware reasoning on multimodal structured data. We explore their ability to reason on tables that integrate both images and text, introducing {\sc {\sc MMTabQA}}, a new dataset designed for this purpose. Our experiments highlight substantial challenges for current AI models in effectively integrating and interpreting multiple text and image inputs, understanding visual context, and comparing visual content across images. These findings establish our dataset as a robust benchmark for advancing AI's comprehension and capabilities in analyzing multimodal structured data.

\end{abstract}

\section{Introduction}

\begin{figure}[t]
\centerline
{\includegraphics[scale=0.17]{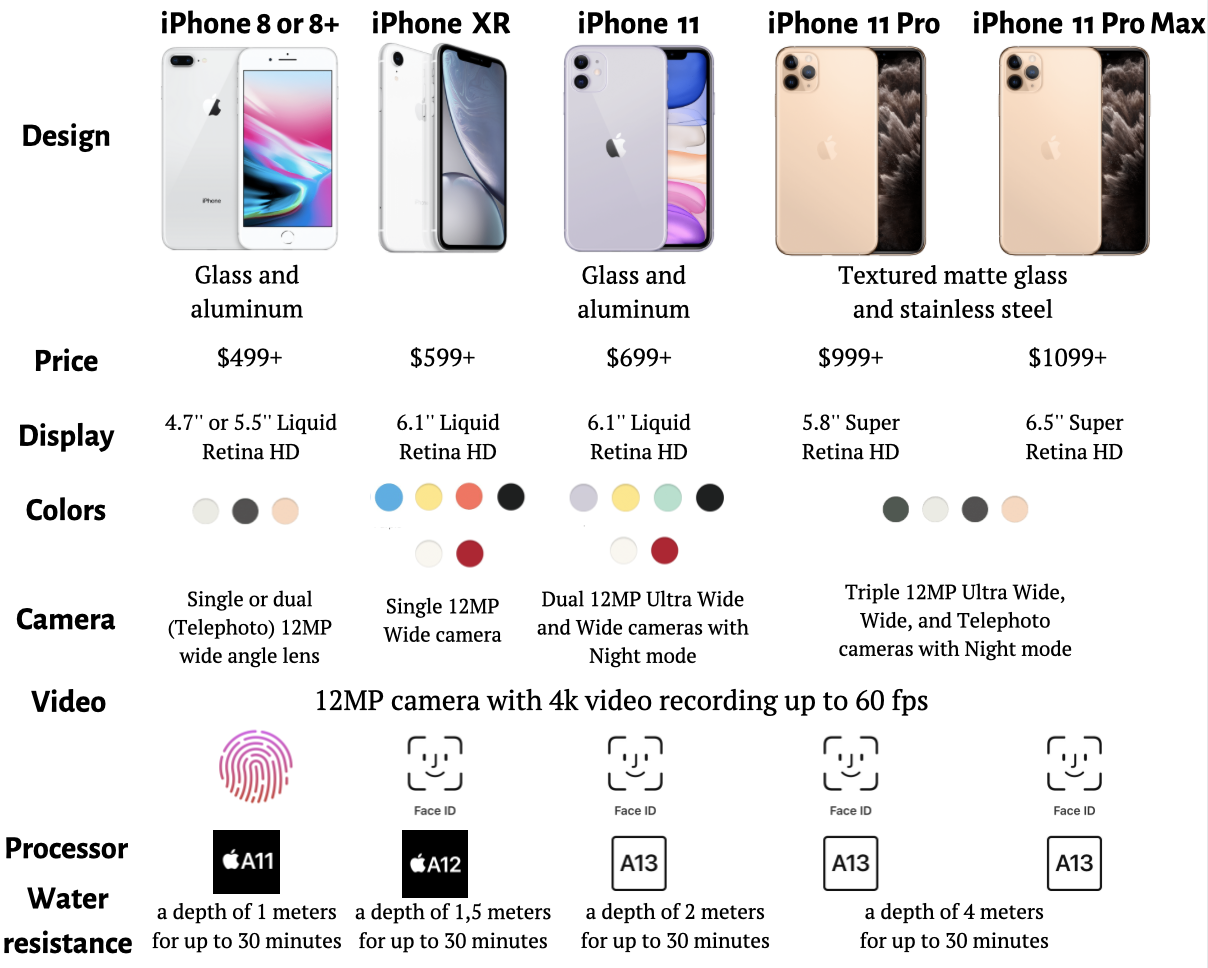}}
\vspace{0.25em}
\small
\begin{tabular}{p{7cm}}
    \noindent \textbf{Q1} How to unlock the phone which has a dual horizontal camera? \\\textbf{A1} Fingerprint Scanner
    
    \vspace{0.25em}
    \textbf{Q2} Which phone combine three camera lens with latest processor? \\\textbf{A2} iPhone 11
    
    \vspace{0.25em}
    \textbf{Q3} Which phone comes with the fewest color options?
    \\\textbf{A3} iPhone 8 or 8+
\end{tabular}
\vspace{-0.5em}
\caption{Multimodal Table comparing iPhone features}
\label{iphone}
\vspace{-1.5em}
\end{figure}

Tables are crucial for efficiently summarizing and conveying information across various fields. In real-world applications, they often include images representing entities, such as team logos in sports scoreboards and product features in E-commerce tables (Fig.~\ref{iphone}). In medicine, tables may display visual symptoms for comparing diseases, while educational tables might include molecular diagrams or images of plant species. Wikipedia tables frequently incorporate images, such as team logos in sports articles or comparative tables for scientists, Nobel laureates, and ship classes. Political party tables often feature election symbols and charts illustrating seat wins. This integration of images enriches the data's depth and informativeness.

% Tables are essential for efficiently summarizing and conveying information across various fields. In real-world applications, they often extend beyond textual data to include images representing entities, such as team logos or flags in tournament scoreboards and product features in E-commerce tables, as shown in Fig.~\ref{iphone}. In medicine, tables might display visual symptoms for comparing diseases, while educational tables could include molecular diagrams or images of plant species. Wikipedia tables frequently incorporate images, such as team logos in sports content or comparative tables for scientists, Nobel laureates, and ship classes. Political party tables often feature election symbols and charts illustrating seat wins. This integration of images enhances the richness and informativeness of the data presented.

Understanding and interpreting these multimodal tables is crucial across various domains. In healthcare, they help doctors compare disease symptoms for accurate diagnosis and treatment planning. In education, students use visual aids in tables to better understand complex concepts. In E-commerce, consumers rely on product comparison tables to make informed purchasing decisions.

Advances in modeling techniques, including table pre-training and targeted fine-tuning, have greatly improved reasoning capabilities for semi-structured tables \cite{muller-etal-2021-tapas, aly-etal-2021-fact}. Furthermore, large language models (LLMs) have shown remarkable performance across diverse domains, achieving state-of-the-art performance on various tabular reasoning tasks \cite{chen-etal-2021-finqa, wang-etal-2021-semeval, lu2023dynamic}. Despite extensive exploration of inference and reasoning over tables most prior works \cite{table_survey} has primarily focus on text-only tables. Thus, developing advanced AI models to process multimodal tables is essential. These models must integrate textual and visual data for comprehensive analysis across fields. Improving these models can enhance the accuracy and efficiency of tasks in healthcare diagnostics, education, and consumer decision-making, thereby enriching data presentation and enhancing its informativeness and utility.

% Enhancing these models can improve the accuracy and efficiency of tasks in healthcare diagnostics, education, and consumer decision-making, enriching data presentation and increasing its informativeness and utility.

% Developing advanced AI models to process multimodal tables is essential. These models must integrate textual and visual data for comprehensive analysis and interpretation across fields. Improving these models can enhance the accuracy and efficiency of tasks in healthcare diagnostics, educational content delivery, and consumer decision-making. This integration enriches data presentation and increases the informativeness and utility of the provided information. 

Reasoning with multimodal tables poses significant challenges. Table reasoning, as illustrated in Fig.~\ref{iphone}, necessitates entity disambiguation within the table context. For instance, disambiguating the A13 square as representing the latest A13 processor chip is essential for answering Q2. Additionally, visual reasoning over individual images in the table is crucial, such as using the phone images to determine the camera alignment for answering Q1. Similarly for Q3, understanding phone colors through their visual representation and counting them could be challenging. To answer these questions well, model must understand the images in the table and its relation to other cells (images and text), which involves complex reasoning such as visual analysis, numerical interpretation, temporal sequencing, and entity relationship identification.

% Answering these questions necessitates understanding multiple images in the table and might also involve complex reasoning, such as visual, numerical, temporal, and entity reasoning.

% However, there are no datasets available to evaluate the performance of Vision-Language models on the complex task of multimodal table reasoning, resulting in this area remaining largely unexplored. 

However, the lack of datasets to evaluate Vision-Language models on multimodal table reasoning has left this area largely unexplored. Therefore, this paper aims to investigate the research question: \emph{Can current Vision-Language models handle complex reasoning in multimodal tables?} To effectively tackle this challenge, we introduce a new task called \textit{knowledge-aware reasoning over multimodal semi-structured tables}. Due to the time-consuming and costly process of curating a new human-annotated dataset on multimodal tables, we repurpose existing Wikipedia datasets into a multimodal format. Our framework replaces recognizable entities in textual Wikipedia tables with their representative images, creating the {\large M}ulti{\large M}odal {\large TAB}le {\large Q}uestion {\large A}nswering ({\sc {\sc MMTabQA}}) dataset repurposed using four Wikipedia tables-based question-answering datasets.

% However, no datasets exist to evaluate the performance of existing Vision-Language models on such complex multimodal table reasoning, leaving this area largely unexplored. Consequently, this paper seeks to address the research question \textit{``Can current Vision-Language models perform complex reasoning over semi-structured multimodal tables?''} 

% In {\sc {\sc MMTabQA}}, we categorize the questions into explicit questions (mentioning an image-replaced entity explicitly), answer-mention questions (referencing an image-replaced entity in the answer), and implicit questions (involving image-replaced entities in intermediate reasoning). Additionally, we create synthetic visual questions by augmenting explicit questions with visual attributes of the mentioned entity. We verify the recast tables and synthetic questions through human evaluation. We evaluate a range of state-of-the-art closed and open-source LLMs and VLMs using diverse modelling approaches on this dataset. Our findings highlight the challenges of entity disambiguation, understanding table structures, and performing visual reasoning within our dataset. 

In {\sc MMTabQA}, we categorize questions into explicit (mentioning an image-replaced entity explicitly), answer-mention (referencing an image-replaced entity in the answer), and implicit (involving image-replaced entities in intermediate reasoning). We also generate synthetic visual questions by enhancing explicit questions with visual attributes of the mentioned entity and validate them through human evaluation. Evaluating various state-of-the-art closed and open-source LLMs and VLMs using diverse modeling approaches on {\sc MMTabQA} reveals challenges in entity disambiguation, understanding table structures, and performing visual reasoning. We aim for our dataset to be a robust benchmark for evaluating Vision-Language Models (VLMs) on complex multimodal tabular reasoning. We summarize our contributions as below:

% Our findings underscore the challenges associated with entity disambiguation, understanding table structures, and performing visual reasoning within our dataset. We aim for this dataset to be a robust benchmark for evaluating VLMs on complex multimodal tabular reasoning.

\begin{itemize}
\vspace{-0.15em}
\setlength{\itemsep}{0pt}
\vspace{-0.25em}
    \item We propose Knowledge-Aware Reasoning over Multimodal Semi-structured Tables and present a framework to repurpose Wikipedia textual-table datasets for multimodal tasks.
\vspace{-0.25em}
    \item Using this framework, we create the {\sc {\sc MMTabQA}} dataset for studying knowledge-aware multimodal reasoning over tables and evaluate various Vision-Language Models (VLMs) using diverse techniques.
\vspace{-0.25em}
    \item Our analysis shows that current VLMs face challenges in performing reasoning on {\sc {\sc MMTabQA}}. They struggle with erroneous entity linking, visual understanding difficulties, and table structure comprehension.
\end{itemize}

\section{Multimodal Tabular Reasoning}
Today's VLMs face multiple challenges when reasoning with multimodal tabular question-answering datasets. Below, we describe these challenges in detail:

%%%%%%%%%%% OLDER CONTENT%%%%%%%%%%%
\subsection{Tabular Multimodal Structure}

Table reasoning is inherently challenging as it needs to rightly interpret semi-structured data, understand complex entity relationships, and integrate diverse contexts \cite{fang2024large}. This difficulty is compounded when processing even a single image with text as an additional modality \cite{de2023visual}. Multimodal table reasoning involves multiple images, while current VLMs are optimized to reason over a single image. Unlike a separated context, these images are semi-structured within the table context. In Fig.~\ref{iphone}, phone images in the table correspond to the named phones in the header, and processor images must be linked to their respective phone columns. Encoding such tables as interleaved multimodal inputs is particularly challenging for VLMs \cite{tian2024mminterleaved}. Thus, exploring various approaches, such as captioning images individually to create a text-only table or representing the entire table as an image, can be further explored.

% Table reasoning is challenging due to the need to interpret semi-structured data accurately, understand complex entity relationships, and integrate diverse contexts. Conversely, processing even a single image as an additional modality proves to be a significant challenge for models. Multimodal Table Reasoning involves multiple images, while many VLMs are tuned to reason over a single image. Additionally, these images are semi-structured within the table context, rather than being unstructured or unrelated. In Fig.~\ref{iphone}, the phone images in the table correspond to the named phone in the header, while the processor images must be linked to their respective phone's column. Such tables should be encoded as interleaved multimodal inputs, posing a challenge for VLMs.  We explore alternative approaches, including captioning images individually to create a text-only table and representing the entire table as a single image for reasoning.

\subsection{Complex Multimodal Reasoning}
Beyond understanding the multimodal table, the model must perform complex reasoning to answer questions. For instance, in Fig.~\ref{iphone}, answering \emph{"Which iPhones have the A13 processor?"} requires the model to identify the A13 chip and link the image of a fingerprint to a fingerprint scanner, generating the corresponding text. This task is more challenging than intermediate reasoning, especially when questions reference image-replaced entities. The table's context, such as comparing iPhone features, is crucial for accurate entity disambiguation. Additionally, answering questions often requires comparing visual attributes across images, demanding a visual understanding beyond simple entity disambiguation. For example, to answer Q1, the model must compare camera placement across all phone images. Moreover, some questions involve reasoning over multiple images, further complicating the task. Answering Q1 requires comparing camera alignment over several phone images, and other questions may require complex reasoning, such as temporal, numerical, and entity reasoning, to derive the correct answer.

\section{{\sc MMTabQA} Dataset}

\label{sec:MMTabQA}

\begin{figure}[t]
    \centering
    \includegraphics[width=0.45\textwidth]{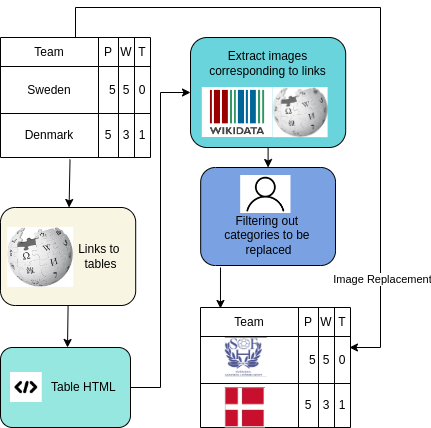}
    \vspace{-0.5em}
    \caption{\small Dataset Creation Pipeline}
    \label{fig:example}
    \vspace{-1.5em}
\end{figure}

\subsection{Original Tabular Dataset}
To diversify the dataset for knowledge-based question answering over multimodal tables, we adapted four existing Wikipedia-based datasets each featuring a variety of real-world entities. Specifically, we adapted the following datasets:
\begin{itemize}
\vspace{-0.3em}
\setlength{\itemsep}{0.25pt}
    \item WikiSQL Dataset \cite{zhong2017seq2sql} to benchmark model capabilities in parsing entities accurately and answering basic SQL-based questions.
    \item WikiTableQuestions dataset \cite{pasupat-liang-2015-compositional} to include questions which require more complex reasoning.
    \item FeTaQA dataset \cite{nan-etal-2022-fetaqa} to include long-form answer based questions which involve multiple row/column reasoning.
    \item HybridQA dataset \cite{chen-etal-2020-hybridqa} which includes extra contextual passages beyond the tables, requiring hybrid complex reasoning.
\vspace{-0.5em}
\end{itemize}

Our dataset and associated code will be released for public use.

\subsection{{\sc MMTabQA} Creation}
% TODO: Reorganize a bit
\subsubsection*{{\sc MMTabQA} Table Generation}
To convert tables from textual to multimodal form, we link textual entity mentions to corresponding images. Wikipedia datasets enable easy access to entity images from Wikipedia Infoboxes or Wikidata entries using Wikipedia page links.

For each dataset, the raw HTML of the table's page is obtained using the corresponding page revisions and the \emph{Jaccard Coefficient} is used to locate the table on the page. The Wikipedia links from the raw HTML table are extracted with their corresponding images, prioritizing Wikipedia Infobox images and using a defined priority order for Wikidata images \cite{lerner2022viquae}. We filter entities for image replacement based on the WikiData P31 "instance of" property and corresponding Wikipedia pageviews to find popular and recognizable images. Over 1,500 unique "instance of" values were annotated for pageview-based filtering across the datasets, while limiting certain values to only use seals, coat of arms, or logo images from the infobox.

% (i) Infobox image from the Wikipedia page  (ii) WikiData property P18 ``image'' (roughly equivalent to the infobox image in Wikipedia articles); (iii) P154 ``logo image"; (iv) P41 ``flag image''; (v) P94 ``coat of arms image''; (vi) P2425 ``service ribbon image''.

Finally we replace linked text in tables with corresponding images from their Wikipedia URLs. In WikiTableQuestions, linked text is directly replaced with scraped images using the original HTMLs provided. For the WikiSQL and FeTaQA datasets, all $(link, text)$ pairs are extracted and filtered from the HTML, and the text in the original table is replaced. In HybridQA, we leverage provided Wikipedia links corresponding to different cells to enhance table quality when replacing text. Additionally, coreference resolution is used to find all mentions of image-replaced entities in their passages in HybridQA, and they are replaced with tags to prevent entity name leaking.

\subsubsection*{{\sc MMTabQA} Question Filtering \& Creation}
We filter out the questions corresponding to the tables of the following types:
% \begin{itemize}

    \noindent \textbf{Explicit Questions} which mention an entity that is replaced by an image in the table.

    \noindent \textbf{Answer-Mention Questions} whose answer contains an entity that is replaced by an image in the table, but the question does not.

    \noindent \textbf{Implicit Questions} where an image-replaced entity is involved in intermediate reasoning but not mentioned in the answer or the question.

% \end{itemize}
We use simple string matching for filtering out the explicit and answer-mention questions and use evidence cells for filtering implicit questions. Finally, we exclude tables lacking a column with at least 30\% images and cap each column at 75\% images to maintain a balance between images and text cells, prioritizing retaining evidence cells. %If the images exceed the limit, we prioritize retaining those essential for reasoning across different questions.
% To remove excess images from the table columns, we replace the images with their original images, keeping the images in the priority order: (i) One explicit mention (question/answer-mention) cell from each question of the table (ii) Explicit mention cells (question/answer-mention) from all questions of the table (iii) At least one cell involving implicit reasoning for every question based on the table (iv) All cells involving implicit reasoning for any questions based on the table (v) Cells which are not involved in any reasoning/mention.
% Based upon the priority described above, we ensure that maximum 75\% of a column contains images. We recheck for explicit, implicit and answer questions and report the final dataset.

We introduce \textbf{visual questions} in our dataset, involving visual aspects of entity images from the table by recasting the explicit mention questions. Visual question creation is limited to specific image categories: landscape collage, logo, seal, flag, coat of arms, and poster. For each image and other images of the same category in the table, VLM (Gemini 1.0 \cite{team2023gemini}) is prompted for category-specific visual attributes. LLM (Gemini) is then prompted to provide a set of unique attributes for the explicit entity's image and to replace its mention in the question with these attributes.

\subsection{{\sc MMTabQA} Validation}
% TODO: Write completely

\paragraph{Tables Validation}
Since we recast existing tables, we first need to verify whether the entity replacements are correct. We sample 250 tables from each data source (total 1000 tables) and have 3 annotators score all unique $(image, original\_text)$ pair per table, verifying the correctness of the image replacements.
% \begin{itemize}
    Label 0 indicates that the image used for the entity is incorrect; %(e.g., the 2001 Championship logo for the 2004 Championship, an invalid image, or an incomprehensible image)
    Label 1 indicates the image represents the entity but is ambiguous for a human to identify; %(e.g., a generic stadium, an F1 racer with an obscured face, a current logo of a previously renamed company, or an unrecognizable town/city collage with generic buildings/images)
    Label 2 indicates the image clearly represents the entity %(e.g., a company logo, a country's flag, or a recognizable monument).
We report the annotation results in Table \ref{tab:agreement}.
% \begin{table}[h!]
% \small
% \centering
% \begin{tabular}{lllll}
% \hline
% \textbf{\begin{tabular}[c]{@{}l@{}}Data\Source\end{tabular}} & \textbf{\begin{tabular}[c]{@{}l@{}}Non-Agreement\\number\end{tabular}} & \textbf{\begin{tabular}[c]{@{}l@{}}0\\number\end{tabular}} & \textbf{\begin{tabular}[c]{@{}l@{}}1\\number\end{tabular}} & \textbf{\begin{tabular}[c]{@{}l@{}}2\\number\end{tabular}} \ \hline
% \textbf{FeTaQA} & \begin{tabular}[c]{@{}l@{}}0.28\\(14)\end{tabular} & \begin{tabular}[c]{@{}l@{}}0.00\\(0)\end{tabular} & \begin{tabular}[c]{@{}l@{}}0.08\\(4)\end{tabular} & \begin{tabular}[c]{@{}l@{}}99.64\\(5030)\end{tabular} \ \hline
% \textbf{HybridQA} & \begin{tabular}[c]{@{}l@{}}0.46\\(28)\end{tabular} & \begin{tabular}[c]{@{}l@{}}0.26\\(16)\end{tabular} & \begin{tabular}[c]{@{}l@{}}2.20\\(134)\end{tabular} & \begin{tabular}[c]{@{}l@{}}97.08\\(5910)\end{tabular} \ \hline
% \textbf{WikiSQL} & \begin{tabular}[c]{@{}l@{}}0.00\\(0)\end{tabular} & \begin{tabular}[c]{@{}l@{}}0.10\\(6)\end{tabular} & \begin{tabular}[c]{@{}l@{}}0.04\\(2)\end{tabular} & \begin{tabular}[c]{@{}l@{}}99.86\\(5688)\end{tabular} \ \hline
% \textbf{\begin{tabular}[c]{@{}l@{}}WikiTable- Questions\end{tabular}} & \begin{tabular}[c]{@{}l@{}}0.43\\(40)\end{tabular} & \begin{tabular}[c]{@{}l@{}}0.26\\(24)\end{tabular} & \begin{tabular}[c]{@{}l@{}}0.13\\(12)\end{tabular} & \begin{tabular}[c]{@{}l@{}}99.19\\(9282)\end{tabular} \ \hline
% \end{tabular}
% \caption{MMTabQA Agreement Statistics}
% \label{tab:agreement}
% \end{table}

\begin{table}[ht!]
\vspace{-0.25em}
\centering
\renewcommand{\arraystretch}{1.3} % Adjust row height

\setlength{\tabcolsep}{2.8pt}
% \begin{tabular}{lllll}
% \hline
% \textbf{\begin{tabular}[c]{@{}l@{}}Data Source\end{tabular}} & \textbf{\begin{tabular}[c]{@{}l@{}}No-\\ Agree-\\ ment\end{tabular}} 
\small
\begin{tabular}{lcccc}
\hline
\bf Data & \bf No  & \textbf{0} & \textbf{1} & \textbf{2} \\
\bf Source  & \bf Agree. & &\\
\hline
\textbf{FeTaQA} & 0.28(14) & 0.00(0) & 0.08(4) & 99.64(5030) \\ 
\textbf{HybridQA} & 0.46 (28) & 0.26 (16) & 2.20 (134) & 97.08 (5910) \\
 % &  &  &  &  \\ 
 % \hline
\textbf{WikiSQL} & 0.00 (0) & 0.10 (6) & 0.04 (2) & 99.86 (5688) \\
 % &  &  &  &  \\ \hline
\textbf{\begin{tabular}[c]{@{}l@{}}WikiTable-\\ Questions\end{tabular}} & 0.43 (40) & 0.26 (24) & 0.13 (12) & 99.19 (9282) \\
 % &  &  &  &  \\ 
 \hline
\end{tabular}
\vspace{-0.5em}
\caption{{\sc MMTabQA} Agreement Statistics: The number in brackets represent absolute number.}
\vspace{-2.0em}
\label{tab:agreement}
\end{table}

\paragraph{Questions Validation}
Explicit, answer-mention, and implicit questions are repurposed from existing tabular QA datasets and do not require additional validation, as the tables themselves are already validated. However, we need to validate the synthetically created visual questions. %We provide the annotators with the multimodal table, the original question, the recast question and the answer and ask them to 
Three annotators score 500 recast questions %from each data source (total 2000 questions) 
as 0 or 1, where 0 indicates that the recast question is incorrect (wrong attribute hallucinated, not uniquely identifiable with table attributes) while 1 indicates appropriate question. We use inter-annotator agreement and obtain 15.6\% questions annotated as 0 while 84.4\% examples are annotated as 1.

% \subsection{{\sc MMTabQA} Dataset}

% \begin{table}[ht]
% \small
% \setlength{\tabcolsep}{2.3pt}
% \centering
% \begin{tabular}{llll}
% \hline
% \textbf{\begin{tabular}[c]{@{}l@{}}Data Source\end{tabular}} &
%   \textbf{\begin{tabular}[c]{@{}l@{}}No. of\\ Questions\end{tabular}} &
%   \textbf{\begin{tabular}[c]{@{}l@{}}No. of\\ Tables\end{tabular}} &
%   \textbf{\begin{tabular}[c]{@{}l@{}}Avg. Img\\ per table\end{tabular}} \\ \hline
% \textbf{WikiSQL}                                                        & 21472 & 9784  & 13.68 \\ 
% % \hline
% \textbf{\begin{tabular}[c]{@{}l@{}}WikiTable Questions\end{tabular}} & 10052 & 1259  & 17.67 \\ 
% % \hline
% \textbf{FeTaQA}                                                         & 7476  & 5898  & 10.43 \\ 
% % \hline
% \textbf{HybridQA}                                                       & 30470 & 8085  & 14.64 \\ 
% % \hline
% \textbf{Overall}                                                        & 69740 & 25026 & 14.10 \\ 
% \hline
% \end{tabular}
% \vspace{-0.5em}
% \caption{\small {\sc MMTabQA} Statistics}
% \label{main-stats}
% \vspace{-1.5em}
% \end{table}
% As described, we create the {\sc MMTabQA} dataset with 69,740 questions over 25,026 tables. Major statistics are in Table \ref{main-stats} and additional statistics are in Appendix \ref{appendix-stats}.
% TODO: Add stats

\subsection{{\sc MMTabQA} Statistics}
As described, we create the {\sc MMTabQA} dataset with 69,740 questions over 25,026 tables. Major statistics are in Table \ref{main-stats}.

\begin{table}[ht]
\small
\centering
\renewcommand{\arraystretch}{1.3} % Adjust row height
\begin{tabular}{llll}
\hline
\textbf{\begin{tabular}[c]{@{}l@{}}Data Source\end{tabular}} &
  \textbf{\begin{tabular}[c]{@{}l@{}}No. of\\ Questions\end{tabular}} &
  \textbf{\begin{tabular}[c]{@{}l@{}}No. of\\ Tables\end{tabular}} &
  \textbf{\begin{tabular}[c]{@{}l@{}}Avg. Img\\ per Table\end{tabular}} \\ \hline
\textbf{WikiSQL}                                                        & 21,472 & 9,784  & 13.68 \\ 
\textbf{\begin{tabular}[c]{@{}l@{}}WikiTable-\\Questions\end{tabular}} & 10,052 & 1,259  & 17.67 \\ 
\textbf{FeTaQA}                                                         & 7,476  & 5,898  & 10.43 \\ 
\textbf{HybridQA}                                                       & 30,470 & 8,085  & 14.64 \\ 
\textbf{Overall}                                                        & 69,740 & 25,026 & 14.10 \\ 
\hline
\end{tabular}
\vspace{-0.5em}
\caption{\small {\sc MMTabQA} Statistics}
\label{main-stats}
\vspace{-1.5em}
\end{table}

% Major statistics are in Table \ref{main-stats} and additional statistics are in Appendix \ref{appendix-stats}.

% \label{appendix-stats}

% \vspace{-0.5em}
Table~\ref{tab:mm_column_stats} highlights the complexity of reasoning involved in different questions based upon the number of columns involved.  We can see FeTaQA involving higher multi-column reasoning questions.

    \begin{table}[ht]
    \centering
    \small
    \setlength{\tabcolsep}{4pt}
    \renewcommand{\arraystretch}{1.3} % Adjust row height

    \begin{tabular}{lll}
    \hline
    \textbf{Data Source} &
      \textbf{\begin{tabular}[c]{@{}l@{}}Single col.\\reasoning\end{tabular}} &
      \textbf{\begin{tabular}[c]{@{}l@{}}Multi col.\\reasoning\end{tabular}} \\ \hline
    \textbf{WikiSQL}            & 17,558 & 3,914 \\ 
    % \hline
    \textbf{\begin{tabular}[c]{@{}l@{}}WikiTable-\\Questions\end{tabular}} & 8,952  & 1,100 \\ 
    % \hline
    \textbf{FeTaQA}             & 4,620  & 2,856 \\ 
    % \hline
    \textbf{HybridQA}           & 26,358 &  4,112 \\ \hline
    \end{tabular}
    \caption{{\sc MMTabQA} Reasoning Complexity}
    \label{tab:mm_column_stats}
\end{table}

% While WikiSQL and WikiTableQuestions balance question types, FeTaQA emphasizes multi-column reasoning. HybridQA stands out for its high proportion of multi-column questions.% , making it a valuable resource for training and evaluating models suited for complex, multimodal reasoning tasks.

Table \ref{tab:question_types} presents the different types of questions in {\sc MMTabQA}, each presenting a different type of challenge. While explicit questions require disambiguating the entity mentioned in the question, answer-mention questions are more complex because they need to generate an image-replaced entity in the answer. Implicit questions on the other hand involve more logical reasoning, while visual questions require the model to understand the visual aspects of images in the table specifically.
% As seen in Table \ref{tab:question_types}, there is a diversity of question types across different datasets, categorized into explicit questions, implicit questions, visual questions, and questions with explicit answer mentions. Further, the HybridQA dataset stands out with a significant number of implicit questions, attributed to passages linked to image entities. On the other hand, WikiSQL and HybridQA contain a substantial proportion of visual questions, underscoring their reliance on visual data for querying. This variety illustrates how different datasets cater to various question types, reflecting the complexity and richness of data sources in contemporary question-answering research.

\begin{table}[ht!]
\small
\centering
\renewcommand{\arraystretch}{1.3} % Adjust row height
\begin{tabular}{lllll}
\hline
\textbf{\begin{tabular}[c]{@{}l@{}}Data\\ Source\end{tabular}} & \textbf{\begin{tabular}[c]{@{}l@{}}Explicit\\ Ques\end{tabular}} & \textbf{\begin{tabular}[c]{@{}l@{}}Implicit\\Ques\end{tabular}} & \textbf{\begin{tabular}[c]{@{}l@{}}Visual\\Ques\end{tabular}} & \textbf{\begin{tabular}[c]{@{}l@{}}Answer-\\Mention \\Ques\end{tabular}} \\ \hline
\textbf{FeTaQA} & 2,499 & 612 & 1,185 & 3,180 \\ 
% \hline
\textbf{\begin{tabular}[c]{@{}l@{}}WikiTable-\\Questions\end{tabular}} & 3,523 & 2,879 & 877 & 2,773 \\ 
% \hline
\textbf{WikiSQL} & 12,956 & 315 & 1,827 & 6,374 \\ 
% \hline
\textbf{HybridQA} & 5,819 & 17,647 & 1,874 & 5,130 \\ 
\hline
\end{tabular}
\vspace{-0.5em}
\caption{Question Type Statistics for {\sc MMTabQA}}
\label{tab:question_types}
\end{table}

\begin{table}[ht!]
\centering
\small
\renewcommand{\arraystretch}{1.3} % Adjust row height
\setlength{\columnsep}{0.1pt}
\begin{tabular}{lcccc}
\hline
\textbf{Domain ($\%$)} & \textbf{WTQ} & \textbf{FetaQA} & \textbf{WikiSQL} & \textbf{HybQA} \\ 
\hline
\textbf{STEM} & {30.18} & {29.00} & {28.87} & {29.28} \\ 
% \hline
\textbf{Media} & 16.04 & 17.73 & 16.49 & 17.75 \\ 
% \hline
\textbf{Biography} & 14.05 & 15.20 & 14.23 & 13.85 \\ 
%\hline
\textbf{None} & 11.83 & 11.40 & 11.38 & 10.92 \\ 
% \hline
\textbf{Europe} & 11.43 & 12.40 & 12.23 & 12.50 \\ 
% \hline
\textbf{NA} & 8.26 & 10.24 & 9.20 & 9.34 \\ 
% \hline
\textbf{P\&G} & 6.98 & 5.91 & 5.96 & 5.36 \\ 
% \hline
\textbf{Technology} & 6.75 & 6.52 & 6.69 & 6.90 \\ 
% \hline
\textbf{Asia} & 6.51 & 7.20 & 6.91 & 6.86 \\ 
% \hline
\textbf{P$\&$R} & 5.24 & - & - & - \\ 
% \hline
\textbf{Literature} & - & 4.83 & 4.63 & 4.81 \\ \hline
\end{tabular}
\caption{Top 10 Domains of the tables categorized based on the topic of the Wikipedia page. P$\&$G: Politics and Government, P$\&$R: Philosophy and Religion, WTQ: WikiTableQuestions, HybQA: HybridQA, NA: North America.}
\label{tab:top_domains}
\end{table}

Table \ref{tab:top_domains} highlights the top 10 domains of the tables in the different datasets, indicating the topics on which the tables and the corresponding questions are based on. Table \ref{tab:statistics-table} provides an overview of image category distributions across four prominent datasets. The analysis reveals a consistent emphasis on human and logo categories across all datasets, indicating these entities are central to the types of questions posed. Beyond humans and logos, there exists notable variability in other categories such as location/landscapes, seals, coat of arms, flags, and posters across the datasets.

Table \ref{tab:entity-distribution-table} shows the distribution of categories of the entities in the answer, highlighting the specific entities based upon which the questions are posed in the dataset. Some additional statistics on the dataset are presented in table \ref{tab:dataset_stats_extra}.

% Please add the following required packages to your document preamble:
% \usepackage{graphicx}
\begin{table*}[!htbp]
\centering
\small
\renewcommand{\arraystretch}{1.3} % Adjust row height

\setlength{\columnsep}{0.1pt}
% \resizebox{\textwidth}{!}{%
\begin{tabular}{lcccccccc}
 \hline
\textbf{Dataset} &
  \multicolumn{1}{l}{\textbf{Human}} &
  \multicolumn{1}{l}{\textbf{\begin{tabular}[c]{@{}l@{}}Location/\\ landscapes\end{tabular}}} &
  \multicolumn{1}{l}{\textbf{Seals}} &
  \multicolumn{1}{l}{\textbf{Coat of Arms}} &
  \multicolumn{1}{l}{\textbf{Flags}} &
  \multicolumn{1}{l}{\textbf{Poster}} &
  \multicolumn{1}{l}{\textbf{Logo}} &
  \multicolumn{1}{l}{\textbf{Miscellaneous}} \\ \hline
\textbf{\begin{tabular}[c]{@{}l@{}}WikiTable-\\Questions\end{tabular}} & 6,305  & 3,082 & 356 & 460  & 831  & 455  & 2,380 & 1,518 \\ 
% \hline
\textbf{FetaQA}             & 10,043 & 3,779 & 478 & 779  & 1,158 & 5,446 & 5,628 & 8,372 \\ 
% \hline
\textbf{WikiSQL}            & 16,915 & 4,518 & 738 & 703  & 1,149 & 751  & 4,572 & 5,856 \\ 
% \hline
\textbf{HybridQA}           & 31,816 & 4,219 & 868 & 2,193 & 2,053 & 2,313 & 7,794 & 11,090 \\ 
 \hline
\end{tabular}%

% }
%\\ \hline
 \vspace{-0.5em}
\caption{Image Category Distribution in {\sc MMTabQA}}
\label{tab:statistics-table}
\end{table*}

% Please add the following required packages to your document preamble:
% \usepackage{graphicx}
\begin{table*}[!htbp]
\centering
\small
\setlength{\columnsep}{0.1pt}
\renewcommand{\arraystretch}{1.3} % Adjust row height

% \resizebox{\textwidth}{!}{%
\begin{tabular}{lcccccccccc}
 \hline
 \small
\textbf{Dataset} &
  \multicolumn{1}{l}{\textbf{Human}} &
  \multicolumn{1}{l}{\textbf{Location}} &
  \multicolumn{1}{l}{\textbf{Product}} &
  \multicolumn{1}{l}{\textbf{Time}} &
  \multicolumn{1}{l}{\textbf{Money}} &
  \multicolumn{1}{l}{\textbf{Event}} &
  \multicolumn{1}{l}{\textbf{Number}} &
  \multicolumn{1}{l}{\textbf{Org.}} &
  \multicolumn{1}{l}{\textbf{Boolean}} &
  \multicolumn{1}{l}{\textbf{Other}} \\ \hline
\textbf{\begin{tabular}[c]{@{}l@{}}WikiTable-\\Questions\end{tabular}} & 1,639  & 1,598 & 628 & 430  & 43  & 117  & 3,947 & 901 & 210 & 539 \\ 
% \hline
\textbf{FetaQA} & 957 & 493 & 1,410 & 1,320  & 33 & 173 & 1,792 & 428 & 99 & 771 \\ 
% \hline
\textbf{WikiSQL} & 3,491 & 3,721 & 649 & 2,026  & 1,866 & 112 & 4,254 & 2,623 & 24 & 4,386 \\ 
% \hline
\textbf{HybridQA} & 4,791 & 5,276 & 2,093 & 4,838 & 296 & 299 & 5,079 & 3,382 & 3 & 4,326 \\ 
 \hline
\end{tabular}%
% }
%\\ \hline
 \vspace{-0.5em}
\caption{Distribution of answer entity categories}
\label{tab:entity-distribution-table}
\end{table*}

\begin{table*}[ht!]
\centering
\small
\renewcommand{\arraystretch}{1.3} % Adjust row height
\begin{tabular}{lccccccc}

\hline
\textbf{Dataset} & \textbf{Avg. No.} & \textbf{Questions} & \textbf{Avg. No.} & \textbf{Total Img} & \textbf{Total Unique} & \textbf{Avg. Unique} & \textbf{Avg. Images} \\
                 & \textbf{of Rows} & \textbf{per Table} & \textbf{of Cols} & \textbf{Occur.} & \textbf{Images} & \textbf{Images} & \textbf{per Table} \\ 
                  \hline
\textbf{\begin{tabular}[c]{@{}l@{}}WikiTable-\\Questions\end{tabular}}       & 18.23 & 7.98 & 6.27 & 32,304  & 16,338  & 17.67 & 25.66 \\ 
% \hline
\textbf{FetaQA}    & 14.44 & 1.27 & 6.12 & 102,785 & 37,238  & 10.43 & 17.43 \\ 
% \hline
\textbf{WikiSQL}   & 13.95 & 2.19 & 6.25 & 207,343 & 37,354  & 13.68 & 21.19 \\ 
% \hline
\textbf{HybridQA}  & 15.87 & 3.77 & 4.50 & 153,246 & 62,346  & 14.64 & 18.95 \\ \hline
\end{tabular}
\caption{Additional {\sc MMTabQA} Statistics. Org. stand for Organization.}
\label{tab:dataset_stats_extra}
\end{table*}

\section{Modelling Strategies} % change name 
\label{sec:modelling_strats}
To benchmark the model performance on our {\sc MMTabQA} dataset, we define four baselines:
\subsection{Partial Input Baseline}
In this baseline, images are excluded, providing only the table with replaced image tags alongside the question to the model. These image tags act as placeholders indicating where images would be in the text format. The model makes guesses about which entities correspond to the image/entity tags for QA. This baseline serves as a lower bound, as models with direct access to images are expected to perform better.

\subsection{Image-captioning Baseline}
Here, the table is converted to a text-only format for reasoning using Language Models (LLMs), using captions generated by VLMs instead of image tags. 
%Like the previous baseline, Question-Answering is conducted on a textual table. However, captions generated from the image modality are also utilized alongside the image/entity tags. This baseline simplifies the problem into two distinct steps.
% \begin{enumerate}
% % \setlength{\itemsep}{0.00pt}
%     \item 
    
- \textbf{Entity Prediction:} Initially, we predict entities for each image occurrence using infobox-style tables. These tables are created with individual rows of the table, where cell corresponding to same column as entity of interest is text-only. VLMs then use this context to predict the original text associated with each image along with a brief visual description of the image.

    % \item 
- \textbf{Question Answering:} After preparing table $T$, question $Q$, and predicted entities $E$ with their visual descriptions $V$, we prompt LLM to generate the answer to $Q$ using $T$ while considering $V$ and $E$, explicitly describing possible inaccuracy of $V$ and $E$.
% \end{enumerate}

Captioning individual images within tables is highly resource-intensive, especially since tables typically contain 10-16 images each. Despite its computational expense, this baseline is valuable for converting the task into text-only format, enabling the use of a larger LLM to handle complex reasoning and integrate visual and textual data effectively.

\subsection{Table-Image Baseline}
Here, we create an image of the table that includes all embedded entity images. This multimodal input, consisting of the table image and textual question, is directly inputted into the model.

\subsection{Interleaved Image-text Baseline}
This baseline fully integrates both visual and textual modalities, providing a comprehensive representation. Unlike the first two baselines, which compromise the visual input, and the Table-Image Baseline, which makes textual reasoning challenging, this model achieves optimal representation by combining both modalities effectively. To encourage prompt understanding, LLMs are employed to perform row pruning on tables before evaluating open-source models.
% TODO: Include the part about pruning table

\subsection{Oracle-Entity Replaced Baseline}
We also evaluate oracle entity-replaced textual tables, which are the original textual tables from which the multimodal tables were derived. We do not report these numbers for visual questions because mere entity replacement is inadequate for addressing such questions. This baseline sets an upper bound for explicit, answer-mention, and implicit questions in our dataset, representing the model's performance when entity disambiguation from the image is perfectly executed for tabular reasoning.

\section{Experiments}
% Please add the following required packages to your document preamble:
% \usepackage{booktabs}
% \usepackage{graphicx}
% \usepackage[table,xcdraw]{xcolor}
% Beamer presentation requires \usepackage{colortbl} instead of \usepackage[table,xcdraw]{xcolor}
\begin{table*}[ht]
\small
\setlength{\tabcolsep}{3.5pt}
\resizebox{\textwidth}{!}{%
\begin{tabular}{@{}lllllllllllll@{}}
\toprule
Dataset &
  \multicolumn{4}{l}{WikiTableQuestions} &
  \multicolumn{4}{c}{WikiSQL} &
  \multicolumn{4}{c}{FetaQA} \\ \midrule
\multicolumn{1}{l|}{Model} &
  \multicolumn{1}{l|}{EQ} &
  \multicolumn{1}{l|}{AQ} &
  \multicolumn{1}{l|}{IQ} &
  \multicolumn{1}{l|}{VQ} &
  \multicolumn{1}{l|}{EQ} &
  \multicolumn{1}{l|}{AQ} &
  \multicolumn{1}{l|}{IQ} &
  \multicolumn{1}{l|}{VQ} &
  \multicolumn{1}{l|}{EQ} &
  \multicolumn{1}{l|}{AQ} &
  \multicolumn{1}{l|}{IQ} &
  VQ \\ \midrule
\multicolumn{13}{c}{Partial Input Baseline} \\ \midrule
\multicolumn{1}{l|}{Gemini-1.5 Flash} &
  40.99 &
  27.38 &
  48.95 &
  \multicolumn{1}{l|}{31.4} &
  39.14 &
  28.71 &
  62.22 &
  \multicolumn{1}{l|}{28} &
  \cellcolor[HTML]{FFCCC9}0.51 &
  0.44 &
  \cellcolor[HTML]{FFCCC9}0.44 &
  \cellcolor[HTML]{FFCCC9}0.47 \\
\multicolumn{1}{l|}{GPT-4o} &
  \cellcolor[HTML]{FFCCC9}57.45 &
  \cellcolor[HTML]{FFCCC9}{\color[HTML]{000000} 38.02} &
  \cellcolor[HTML]{FFCCC9}70.83 &
  \multicolumn{1}{l|}{\cellcolor[HTML]{FFCCC9}42.40} &
  \cellcolor[HTML]{FFCCC9}52.57 &
  \cellcolor[HTML]{FFCCC9}43.86 &
  \cellcolor[HTML]{FFCCC9}72.38 &
  \multicolumn{1}{l|}{\cellcolor[HTML]{FFCCC9}{\color[HTML]{000000} 39.00}} &
  0.51 &
  \cellcolor[HTML]{FFCCC9}0.46 &
  0.42 &
  0.44 \\
\multicolumn{1}{l|}{Llama-3 70B} &
  41.13 &
  26.48 &
  43.75 &
  \multicolumn{1}{l|}{31.8} &
  41.117 &
  30.75 &
  61.27 &
  \multicolumn{1}{l|}{30.6} &
  0.52 &
  0.46 &
  0.45 &
  0.48 \\
\multicolumn{1}{l|}{Mixtral 8x7B} &
  26.56 &
  9.90 &
  30.26 &
  \multicolumn{1}{l|}{20.2} &
  23.42 &
  17.71 &
  28.88 &
  \multicolumn{1}{l|}{19.2} &
  0.44 &
  0.39 &
  0.38 &
  0.39 \\ \midrule
\multicolumn{13}{c}{Oracle-Entity Replaced Baseline} \\ \midrule
\multicolumn{1}{l|}{Gemini-1.5 Flash} &
  74.89 &
  78.19 &
  54.86 &
  \multicolumn{1}{l|}{-} &
  82.28 &
  81.86 &
  77.46 &
  \multicolumn{1}{l|}{-} &
  \cellcolor[HTML]{FFCCC9}0.56 &
  \cellcolor[HTML]{FFCCC9}0.50 &
  0.41 &
  - \\
\multicolumn{1}{l|}{GPT-4o} &
  \cellcolor[HTML]{FFCCC9}87.80 &
  \cellcolor[HTML]{FFCCC9}{\color[HTML]{000000} 84.86} &
  \cellcolor[HTML]{FFCCC9}84.55 &
  \multicolumn{1}{l|}{-} &
  \cellcolor[HTML]{FFCCC9}85.57 &
  \cellcolor[HTML]{FFCCC9}82.71 &
  \cellcolor[HTML]{FFCCC9}79.05 &
  \multicolumn{1}{l|}{39.00} &
  0.53 &
  0.48 &
  \cellcolor[HTML]{FFCCC9}0.43 &
  - \\
\multicolumn{1}{l|}{Llama-3 70B} &
  75.74 &
  75.31 &
  58.85 &
  \multicolumn{1}{l|}{-} &
  78.28 &
  78.57 &
  68.25 &
  \multicolumn{1}{l|}{-} &
  0.49 &
  0.46 &
  0.41 &
  - \\
\multicolumn{1}{l|}{Mixtral 8x7B} &
  54.89 &
  53.87 &
  40.69 &
  \multicolumn{1}{l|}{-} &
  59.28 &
  69.28 &
  33.96 &
  \multicolumn{1}{l|}{-} &
  0.44 &
  0.41 &
  0.33 &
  - \\ \midrule
\multicolumn{13}{c}{Image-Captioning Baseline} \\ \midrule
\multicolumn{1}{l|}{Gemini-1.5 Flash} &
  52.34 &
  42.16 &
  51.39 &
  \multicolumn{1}{l|}{42.2} &
  50.42 &
  40.85 &
  67.30 &
  \multicolumn{1}{l|}{46.6} &
  0.57 &
  0.46 &
  0.42 &
  0.43 \\ \midrule
\multicolumn{13}{c}{Table-as-an-Image Baseline} \\ \midrule
\multicolumn{1}{l|}{Gemini-1.5 Flash} &
  44.22 &
  25.65 &
  41.01 &
  \multicolumn{1}{l|}{37.8} &
  47.08 &
  35.75 &
  52.38 &
  \multicolumn{1}{l|}{35.25} &
  0.62 &
  0.43 &
  0.42 &
  0.47 \\
\multicolumn{1}{l|}{GPT-4o} &
  \cellcolor[HTML]{FFCCC9}64.6 &
  \cellcolor[HTML]{FFCCC9}39.60 &
  \cellcolor[HTML]{FFCCC9}67.00 &
  \multicolumn{1}{l|}{\cellcolor[HTML]{FFCCC9}51.8} &
  \cellcolor[HTML]{FFCCC9}55 &
  \cellcolor[HTML]{FFCCC9}43.20 &
  \cellcolor[HTML]{FFCCC9}62.22 &
  \multicolumn{1}{l|}{\cellcolor[HTML]{FFCCC9}54.4} &
  \cellcolor[HTML]{FFCCC9}0.65 &
  \cellcolor[HTML]{FFCCC9}0.47 &
  \cellcolor[HTML]{FFCCC9} &
  \cellcolor[HTML]{FFCCC9}0.49 \\
\multicolumn{1}{l|}{Qwen-VL-chat} &
  14.04 &
  4.51 &
  9.375 &
  \multicolumn{1}{l|}{12} &
  9.58 &
  7.14 &
  35.23 &
  \multicolumn{1}{l|}{8.4} &
  0.49 &
  0.33 &
  0.31 &
  0.36 \\
\multicolumn{1}{l|}{CogAgent-VQA} &
  14.89 &
  5.95 &
  11.28 &
  \multicolumn{1}{l|}{9.4} &
  13.07 &
  11.52 &
  19.36 &
  \multicolumn{1}{l|}{8.8} &
  0.45 &
  0.29 &
  0.15 &
  0.11 \\
\multicolumn{1}{l|}{Intern-VLM-4khd} &
  26.67 &
  13.87 &
  22.22 &
  \multicolumn{1}{l|}{17.2} &
  28.71 &
  18 &
  29.84 &
  \multicolumn{1}{l|}{9.6} &
  0.52 &
  0.36 &
  0.32 &
  0.34 \\ \midrule
\multicolumn{13}{c}{Interleaved Image-text Baseline} \\ \midrule
\multicolumn{1}{l|}{Gemini-1.5 Flash} &
  60.42 &
  33.33 &
  50.44 &
  \multicolumn{1}{l|}{50.39} &
  53.22 &
  40.17 &
  \cellcolor[HTML]{FFCCC9}62.90 &
  \multicolumn{1}{l|}{48.02} &
  0.52 &
  0.42 &
  0.42 &
  \cellcolor[HTML]{FFCCC9}{\color[HTML]{333333} 0.51} \\
\multicolumn{1}{l|}{GPT-4o} &
  \cellcolor[HTML]{FFCCC9}72.47 &
  \cellcolor[HTML]{FFCCC9}49.26 &
  \cellcolor[HTML]{FFCCC9}69.6 &
  \multicolumn{1}{l|}{\cellcolor[HTML]{FFCCC9}47.6} &
  \cellcolor[HTML]{FFCCC9}66.5 &
  \cellcolor[HTML]{FFCCC9}48.93 &
  57.77 &
  \multicolumn{1}{l|}{\cellcolor[HTML]{FFCCC9}{\color[HTML]{000000} 54}} &
  \cellcolor[HTML]{FFCCC9}0.56 &
  \cellcolor[HTML]{FFCCC9}0.51 &
  \cellcolor[HTML]{FFCCC9}0.46 &
  0.49 \\
\multicolumn{1}{l|}{Qwen-VL-chat} &
  12.86 &
  6.64 &
  11.61 &
  \multicolumn{1}{l|}{10.29} &
  9.59 &
  5.38 &
  12.88 &
  \multicolumn{1}{l|}{7.09} &
  0.16 &
  0.17 &
  0.05 &
  0.09 \\
\multicolumn{1}{l|}{Idefics-Mantis} &
  10.46 &
  2.62 &
  10.39 &
  \multicolumn{1}{l|}{8.49} &
  2.8 &
  5.69 &
  9.09 &
  \multicolumn{1}{l|}{3.61} &
  0.34 &
  0.22 &
  0.30 &
  0.3 \\ \bottomrule
\end{tabular}%
}
%\vspace{-0.75em}
\caption{Results on sampled subset of MMTabQA. Substring match is reported for Wiki-realted data sources and ROUGE-L is reported for FetaQA data source. EQ - Explicit Questions, AQ - Answer-Mention Questions, IQ - Implicit Questions, VQ - Visual Questions. Best performing models are highlighted in red.}
\label{tab:my-table}
%\vspace{-0.75em}
\end{table*}

\paragraph{Models used} %We use a mix of open-source and closed-source models to benchmark performance on our dataset. We use Google's closed-source Gemini 1.5 Flash for all five modeling approaches. We use open-source textual LLMs, specifically LLaMa-3 70B \cite{touvron2023llama} and Mixtral 8x7B \cite{jiang2024mixtral}, for the textual Oracle-Entity Replacement baseline and the Partial Input baseline. For table-as-image and interleaved baseline, we utilize GPT-4o by OpenAI \cite{achiam2023gpt} and Qwen-VL Chat \cite{bai2023qwen}. Further, we also benchmark CogAgent-VQA \cite{hong2023cogagent} and Intern-VLM-xComposer-4khd \cite{chen2024far} for the table-image baseline. Additionally, we benchmark the Idefics-Mantis \cite{jiang2024mantis} model, which is specifically trained for handling multiple interleaved images for the interleaved baseline. Note that due to computational constraints with Table Captioning baseline we only run Gemini 1.5 Flash on it.

%We evaluated our models using a combined approach of few-shot learning \cite{brown2020language} and Chain of Thought (COT) prompting \cite{wei2023chainofthought}. COT helps the model understand the reasoning process for a question, while few-shot examples provide insight into the answer format. We used either 4 or 8 examples for prompting, depending on the model and resources available. Sample prompts are provided in the appendix. % merge model & prompting, Methodology/ Approaches
We employ a combination of open-source and closed-source models to benchmark performance on our dataset. Specifically, Google's closed-source Gemini 1.5 Flash \& Open AI's GPT-4o \cite{achiam2023gpt} are utilized for the majority of modeling approaches. For the textual Oracle-Entity Replacement baseline and the Partial Input baseline, open-source textual LLMs, namely LLaMa-3 70B \cite{touvron2023llama} and Mixtral 8x7B \cite{jiang2024mixtral}, are employed. For the table-as-image and interleaved baselines, open-sourced Qwen-VL Chat \cite{bai2023qwen} is used. Additionally, CogAgent-VQA \cite{hong2023cogagent} and Intern-VLM-xComposer-4khd \cite{chen2024far} are benchmarked for the table-image baseline, and the Idefics-Mantis model \cite{jiang2024mantis}, specifically trained for handling multiple interleaved images, is used for the interleaved baseline. Due to computational constraints, only Gemini 1.5 Flash is run on the Table Captioning baseline.

Our prompting methodology combines few-shot learning \cite{brown2020language} and Chain of Thought (COT) prompting \cite{wei2023chainofthought}. COT aids the model in understanding the reasoning process behind a question, while few-shot examples guide the expected answer format. Depending on the model and available resources, we use either 4 or 8 examples for prompting. Sample prompts can be found in the appendix.
\paragraph{Evaluation Metrics} We used different metrics for the two types of answers in our tasks. For single-word or phrase answers, like those in WikiTableQuestions and WikiSQL, we used Substring Match, which checks for the correct answer within the predicted text. For long-form or sentence answers, like those in FeTaQA, we used ROUGE-L \cite{lin-2004-rouge}, which evaluates the longest common subsequence between predicted and reference texts. Detailed evaluations for each data source and question type, with multiple metrics, are in the appendix. %section mention

% Mention answer pruning briefly
% Mention that we report results for each q-type separately

\paragraph{Evaluation Benchmark} We sample out 20\% questions per dataest per question type, sampling at least 500 questions and maximum of 700 questions for our test set. %Due to resource constraints with GPT-4o, we perform stratified sampling using the type of Wikipedia page to limit the test-set to 500 questions specifically.

\section{Result and Analysis}
\label{sec:results}
We observe that our models' performance varies significantly across different datasets, methods, and models. We provide a detailed analysis of these variations below: %(All numbers reported for comparison are from WikiTableQuestions explicit):

\subsection{Performance across Strategies}
% Lower bound upper bound exists -> Also observe that lowest/highest scores.
% As expected, table-as-image has multimodal info aa rahi, performs better than lower bound.
% However, interpreting table structure is hard from just an image, and therefore its performance is still little low. (numbers)
% Interleaved mein proper text and images fully encoded -> much better -> best perf -> still lower than upper bound (numbers)
% Another approach tried -> image as caption -> encodes multimodal information but not as good as interleaved -> performance between table-as-iamge and interleaved (numbers). Again shows interpreting table structure from image is hard.

As described in Section \ref{sec:modelling_strats}, the Partial Input Baseline forms the lower bound for our experiments, as observed in Table \ref{tab:my-table}. Moreover, the Oracle-Entity Replaced Baseline establishes the experimental upper bound, showcasing superior performance compared to all other baselines. This baseline reflects the model's performance under ideal conditions where entity disambiguation is executed with 100\% accuracy.

The Table-as-image baseline integrates missing image information beyond what the Partial Input baseline provides, resulting in an expected improvement in performance. However, the challenge of interpreting table structure directly from the image remains evident for LLMs, thus leading to consistent performance limitations.

We extend upon this with the interleaved Image-text baseline, employing separate encoding for text and images, which yields enhanced representations and improved performance relative to the Table-as-image baseline. However, it is noted that this performance does not achieve parity with our defined Upper Bound, highlighting avenues for further enhancement.

We additionally explore the image-as-caption approach, aiming to encode multimodal information into textual form for question answering. Our findings indicate that while this method is less effective than the interleaved Image-text baseline, it outperforms the table-as-image baseline. This underscores the persistent challenge of accurately interpreting table structure and text from a singular image.

\subsection{Performance across Models} 
Closed-source models like GPT-4o and Gemini-1.5 Flash outperform open-source models in multimodal tasks due to advanced training techniques and better integration of visual and textual data. In text-only tasks, the performance gap between open-source and closed-source models narrows significantly, with open-source models like Llama-3 providing competitive results.

Overall, we see that closed-source models generally outperform open-source models. GPT-4o demonstrates the best performance, achieving a substring match as high as 60.6\% and 42.6\% for WikiTableQuestion Explicit and Answer-mention Questions in the interleaved images approach. This is closely followed by Gemini-1.5 Flash, which achieves a substring match of up to 61\% and 33.33\% for the same dataset subsection. We also note that the true reasoning capabilities of GPT-4o might be more advanced when provided with Interleaved input, as it refuses to answer some questions due to policy violations.

Notably, open-source models provide competitive results in text-only baselines. Here, the performance gap between is around 10\% to 20\%, which indicates that open-source textual LLMs with a large number of parameters are competitive with state-of-the-art closed-source textual LLMs. While the performance of Llama-3 is on-par, the performance of Mixtral 8x7B lags behind. This is because with ~9X more parameters, Llama-3 is capable of much more complex reasoning and parametric knowledge than Mixtral8x7B. 
Furthermore, the Partial Input baseline demonstrates that Open Source models leverage real-world knowledge to infer missing entities.

Their performance notably declines in multimodal baselines, particularly in approaches like Table as an Image and Interleaved Text-Image. In Vision-Language models, the disparity between Open-Source and Closed-Source models becomes more pronounced. Table-as-image models encounter challenges such as entity disambiguation within tables, highlighting deficiencies in parametric multimodal knowledge and table structure parsing, which reflects their relatively weaker Vision Encoders. Similarly, in interleaved models, Open-Source counterparts struggle to contextualize multiple images, often resulting in nonsensical answers influenced predominantly by one image rather than considering all provided images.

% \subsection{Open Source v/s Closed Source Models}
% \label{sec:oss vs close}
\subsection{Performance across Data Sources}
We examine Table~\ref{tab:my-table} to gain deeper insights into our proposed tasks and model performance.  We observe that the performance of models is similar on the WikiTableQuestions dataset and the WikiSQL datasets, since both are short-form question datasets and require a similar kind of entity disambiguation as a challenge for the question-answering. On FeTaQA, we notice that the ROUGE-L scores themselves don't vary much between Upper Bound and Lower Bound. This is because majority of the N-grams used for computing the metric wouldn't involve the image-replaced entity. We observe a significant decrease in scores of Image-Captioning baseline on FeTaQA dataset. This decline is likely due to the inclusion of text from provided captions and visual descriptions, which adversely affected recall on the gold summaries.  In addition, we observe slight variations in ROUGE scores, which offer only a rough indication of VLM performance.

 \begin{figure*}[ht]
    \centering
     \vspace{-0.5em}
    \begin{subfigure}
        \centering
        \includegraphics[height=1.82in]{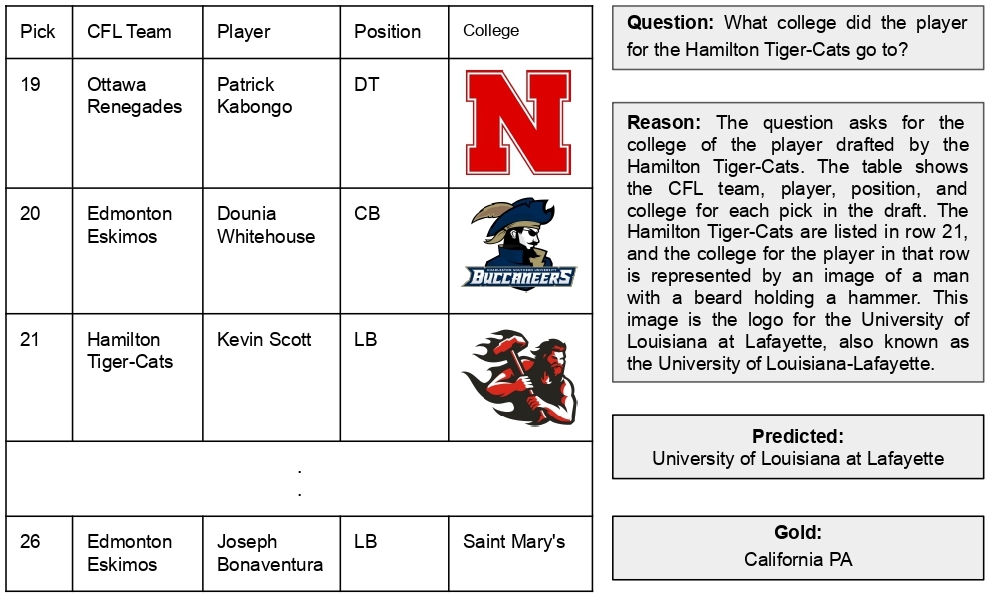}
        \label{fig:left1}
    \end{subfigure}
     \vspace{-0.5em}
    \begin{subfigure}
        \centering
        \includegraphics[height=1.8in]{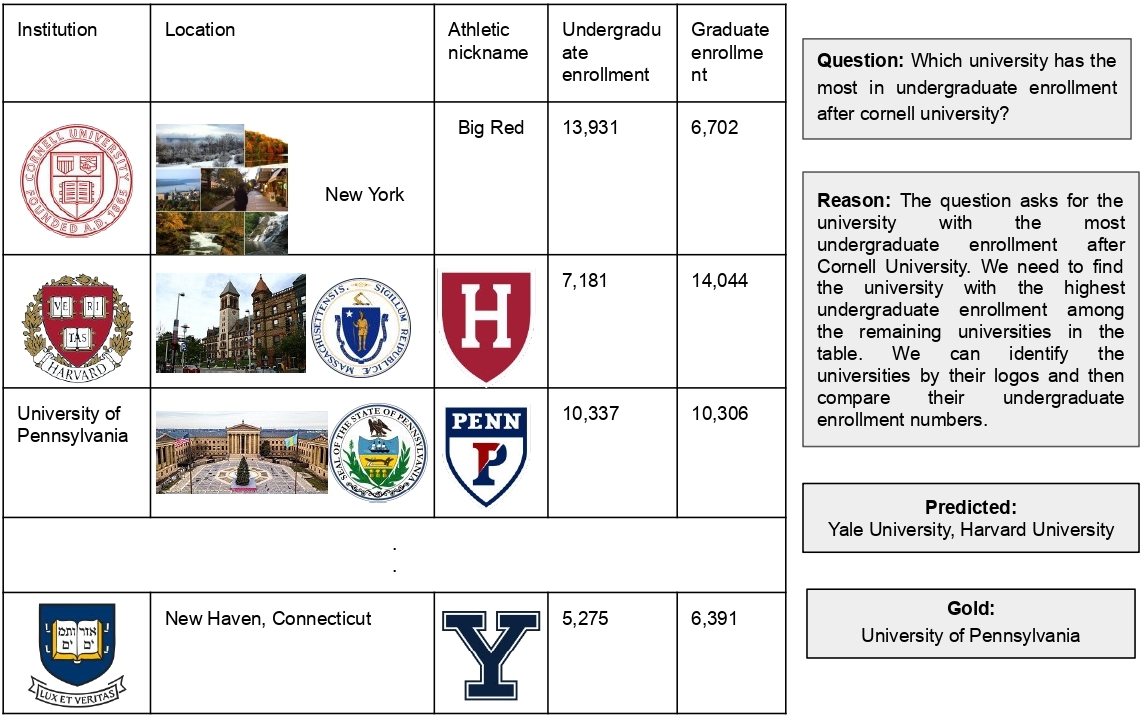}
        \label{fig:right1}
    \end{subfigure}
    \vskip\baselineskip
     \vspace{-0.5em}
    \begin{subfigure}
        \centering
        \includegraphics[height=1.7in]{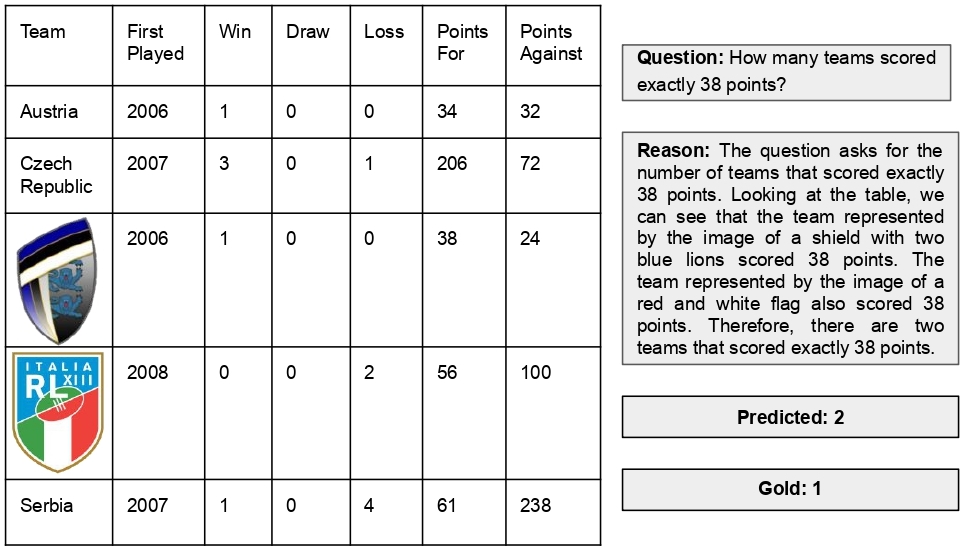}
        \label{fig:left2}
    \end{subfigure}
    \vspace{-0.5em}
    \begin{subfigure}
        \centering
        \includegraphics[height=1.7in]{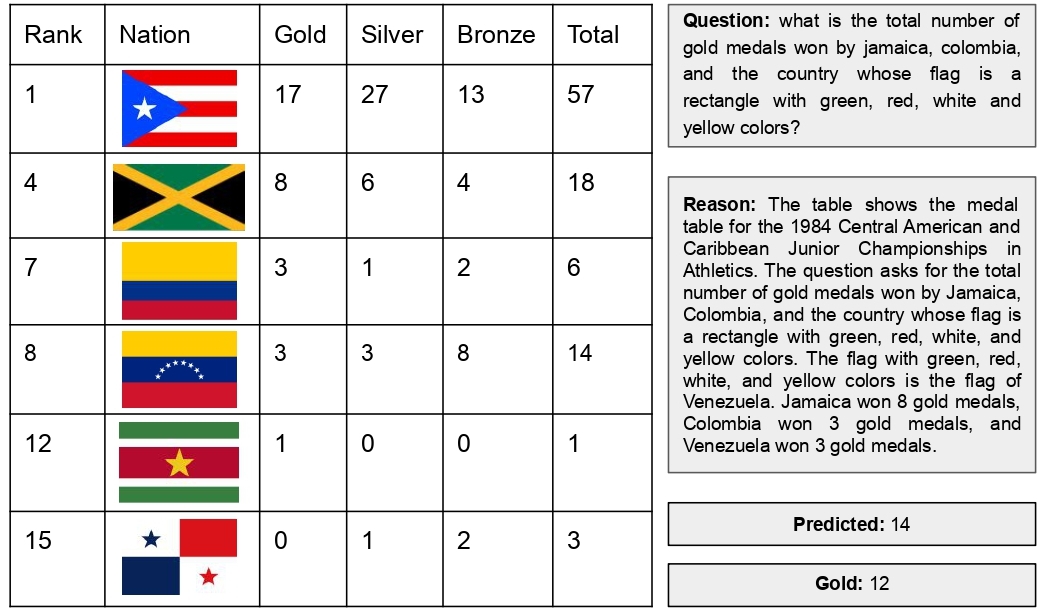}
        \label{fig:right2}
    \end{subfigure}
    
    \vspace{-0.5em}
    \caption{Clockwise from left - (a): Table about College Football, (b): Table about College Enrollment, (c): Table about 1984 Central American Games, (d): Table about International Football.}
    \label{fig:subfigures}
    \vspace{-1.00em}
\end{figure*}

\subsection{Performance across Question types}
Models typically exhibit superior performance in scenarios involving simple reasoning. Conversely, tasks requiring complex reasoning or multi-step inference frequently lead to model failures. This observed trend underscores the challenges faced by current models in handling intricate reasoning processes. Explicit questions, as defined in the preceding discussion, contain clear and specific entity mentions within the query, facilitating their resolution by computational models. This assertion is substantiated by the data presented in Table \ref{tab:my-table}, where explicit questions achieve the highest parameter scores in comparison to other question types.

Implicit questions, in contrast, necessitate additional reasoning to infer the answers, resulting in lower model performance relative to explicit questions. When analyzing the WikiSQL dataset, implicit questions achieve a higher performance metric (52.3\%) compared to explicit questions (47.08\%). This discrepancy can be attributed to the nature of reasoning required by each question type. For tasks requiring very complex reasoning, as seen in WikiTableQuestions (WTQ), explicit questions tend to perform better. However, in the context of WikiSQL, where reasoning is primarily simple and SQL-based, implicit questions exhibit superior performance. This trend is corroborated by the oracle-baseline performance observed on both WikiSQL and WTQ datasets.

Explicit Answer Mention Questions pose a significant challenge as they require the generation of answers that mention entities replaced by images. This task demands precise entity disambiguation, leading to lower performance metrics across all data sources. Specifically, analyzing Gemini's performance in the "Table as an Image" approach, there is a marked decrease in performance—over 10\%—for answer mention questions compared to other question types. 

Visual questions perform better than answer-mention questions but underscore a significant limitation of current language models. These questions necessitate a visual understanding of entities depicted in images, a task that is considerably more challenging than leveraging images for explicit or implicit reasoning. This challenge is reflected in the performance metrics: 37.8\% on WikiSQL and 35.25\% on WikiTableQuestions using the "Table as an Image" approach. While the FeTaQA dataset presents valuable information, its current metrics limit the extent of inferences we can draw.

\section{What did we learn?}
A significant observation from our experiments is the varied performance of models across different baseline settings and the reasons underlying these disparities. GPT-4o demonstrates competitive performance across all data sources and methodologies and those results are used for our analysis. We list some primary issues below:

 \paragraph{ Entity Disambiguation}: The model inaccurately identifies entities from images, leading to errors. For instance, in Figure 3(a), the model misidentifies the logo of California PA as that of the University of Lafayette. 

 \paragraph{ Identification of Visual Attributes}: This presents challenges for multimodal models, particularly in recognizing crucial visual elements within an image required to answer associated questions. For example, in Figure 3(d), the model’s failure to correctly identify a flag based on its colors results in inaccurate responses.

\paragraph{ Handling Excessive Content}: Challenges arise in handling excessive content, leading to instances of incomplete or incorrect retrieval by the model. Figure 3(c) illustrates such a scenario where the model’s incomplete comprehension of the table leads to erroneous conclusions.

\paragraph{ Incorrect Conclusions}: Despite correctly identifying entities at times, the model occasionally reaches incorrect conclusions, possibly due to incorrect reasoning, resulting in erroneous answers, as depicted in Figure 3(b).
These findings highlight the weaknesses of VLMs in image handling, particularly concerning the capabilities of their vision encoders. Notably, open-source models underperform compared to closed-source models as image complexity increases, both in intricacies (e.g., the table-as-an-image approach) and in quantity (e.g., the interleaved-image approach).

Open-source VLMs often lack vision encoders capable of handling intricate or multiple images, resulting in inaccurate interpretations. Moreover, even when some level of image interpretation is achieved, the limited reasoning abilities of these models render them highly ineffective for comprehensive analysis.

Additionally, we also perform a quantitative analysis of these errors on 720 randomly sampled incorrect responses of GPT-4o based on the Table-as-image and Interleaved input approaches on broader error categories. The quantitative insights are given below:
% Please add the following required packages to your document preamble:
% \usepackage{booktabs}
% \usepackage{graphicx}
\begin{table}[!htbp]
\centering
\small
\renewcommand{\arraystretch}{1.3} % Adjust row height

% \resizebox{\columnwidth}{!}{%
\begin{tabular}{@{}llll@{}}
\hline
\bf Type  &\bf WikiSQL &\bf WikiTable &\bf FetaQA \\ \hline
1 & 32.17\%  & 34.48\%    & 22.08\% \\
2 & 3.04\%   & 2.58\%     & 0.83\%  \\
3 & 54.34\%  & 44.82\%    & 63.75\% \\
4 & 10.43\%  & 18.10\%    & 13.33\% \\ \bottomrule
\end{tabular}%
% }
\caption{Error Analysis - A dataset perspective. Type 1: Entity Disambiguation Issues, Type 2: Context Length Related Issues, Type 3: Reasoning and Text Input Errors, Type 4: Identification of Visual Attributes. }
\label{tab:datasetwise}
\end{table}

% Please add the following required packages to your document preamble:
% \usepackage{booktabs}
% \usepackage{graphicx}
\begin{table}[!htbp]
\centering
\renewcommand{\arraystretch}{1.3} % Adjust row height
\small

\begin{tabular}{@{}lllll@{}}
\hline
\bf Type  &\bf EQ &\bf AQ &\bf IQ &\bf VQ \\ \midrule
1 & 28.65\%        & 43.57\%      & 30.81\%        & 15.00\%      \\
2 & 1.75\%         & 0.55\%       & 2.32\%         & 3.88\%       \\
3 & 64.91\%        & 48.04\%      & 61.05\%        & 44.44\%      \\
4 & 4.67\%         & 7.82\%       & 5.81\%         & 36.67\%       \\ \hline
\end{tabular}%
% }
\caption{Error Analysis - A question type perspective. Type 1: Entity Disambiguation Issues, Type 2: Context Length Related Issues, Type 3: Reasoning and Text Input Errors, Type 4: Identification of Visual Attributes.}
\label{tab:questionwise}
\end{table}
Our label classification is as follows:
\begin{enumerate}
\item Entity Disambiguation Issues: Instances where the model fails to accurately identify the entity mentioned in the question, leading to incorrect interpretations (Fig 3a).
\item Context Length-Related Issues: Cases where the model struggles to comprehend prompts due to lengthy context or multiple images, resulting in incorrect or no output.
\item Reasoning and Text Input Errors: Situations where the model's final output is incorrect due to faulty table interpretation, erroneous information extraction (Fig 3c), model hallucination, or incorrect reasoning (Fig 3b).
\item Visual Attribute Identification Errors: Instances where the model incorrectly identifies visual aspects of an image, leading to erroneous answers (Fig 3d).
\end{enumerate}

 Our analysis reveals that the majority of errors can be attributed to reasoning errors, entity disambiguation issues, and difficulties with visual aspect identification, in that order. These findings highlight the current models' inability to effectively process multimodal table data for QA purposes, thereby reinforcing the necessity of our dataset.

\section{Comparison with Related Work}

Recent advancements in natural language processing (NLP) have expanded beyond traditional homogeneous tables to incorporate additional modalities. Works such as \cite{chen-etal-2020-hybridqa, zhu-etal-2021-tat, chen-etal-2021-finqa, zhao-etal-2022-multihiertt} integrate paragraph context alongside tables for enhanced tabular question answering. Meanwhile, efforts like \cite{Talmor2021MultiModalQACQ, li-etal-2022-mmcoqa} introduce images alongside text and tables, yet they do not address the non-homogeneous modalities found in {\sc MMTabQA} tables, containing text-only tables. 

Visual Table understanding has also gained attention, with approaches such as \cite{zheng2024multimodal, kim2024tablevqa} converting textual tables into visual formats for multimodal reasoning. However, these efforts do not capture the multimodal complexities inherent in {\sc MMTabQA}. Datasets like \cite{okvqa, aokvqa} focus on general knowledge for Visual Question Answering, while others \cite{lerner2022viquae, encyclopedic_vqa, Chen2023CanPV} emphasize fine-grained entity knowledge within images. Multimodal Entity Linking and Disambiguation across different modalities are explored in \cite{moon-etal-2018-multimodal-named, wang-etal-2022-wikidiverse}, echoing the entity linking challenges posed by {\sc MMTabQA}. In Visual Question Answering, advancements have been made in handling multiple images \cite{10.24963/ijcai.2023/146, jiang2024mantis}, though it remains less explored compared to single-image tasks.

Unlike prior work, which typically focuses on either textual or visual elements separately, our task confronts the novel challenge of multimodal tables that integrate multiple texts and multiple images within table cells. This involves addressing explicit, implicit, and answer-mention questions, while also advancing visual understanding within the framework of table-based reasoning.

\section{Conclusion}

This research explores whether NLP models can effectively reason with knowledge on multimodal structured data. We investigate their ability to process tables that combine images and text, introducing {\sc MMTabQA}, a new dataset for this task. Our experiments reveal substantial challenges for AI models in integrating and interpreting multiple text and image inputs, understanding visual context, and comparing visual content across images. Our findings position {\sc MMTabQA} as a crucial benchmark for advancing AI's capabilities in analyzing multimodal structured data.

\textbf{Future Directions.}  Our research presents opportunities for expansion by enhancing existing Wikipedia-derived datasets through augmentation and proposing a human-annotated dataset from real-world multimodal tables beyond Wikipedia. Diversifying with additional datasets will enrich our dataset's diversity and scope. Addressing model errors during retrieval is a significant challenge, tackled through Retrieval-Augmented Generation (RAG). Moreover, optimizing open-source models tailored to our task is crucial, focusing on efficient models capable of achieving results comparable to computationally intensive counterparts. These efforts aim to advance multimodal table reasoning.

\section*{Limitations}
%Several limitations are of interest in our study. Firstly, due to budgetary and computational constraints, not all models were fully optimized, possibly leading to underestimated performance outcomes. Secondly, the linguistic constraints of our study, particularly its reliance on English for developing multimodal reasoning datasets and methodologies, emphasize the imperative for linguistic diversity in NLP applications to ensure inclusivity and broader applicability. Given the novelty of our investigation, it is essential to acknowledge that our findings may not encompass all scenarios, indicating the necessity for further exploration.%
Our work has several notable limitations. Chiefly, financial and computational resource constraints prevented us from fine-tuning all the models considered, potentially underrepresenting their capabilities beyond our primary focus. Additionally, the language limitations in this research, particularly the emphasis on English for creating Multimodal Reasoning datasets and methodologies, highlight the necessity of linguistic diversity in NLP applications to ensure broader applicability and inclusivity. Considering the novelty of the task, it is also important to recognize that our insights may not be exhaustive, pointing to the potential for future research.

\section*{Ethical Statement}
As the work's authors, we certify that our investigation and publication adhere to the strictest ethical guidelines.
For the purpose of making our results more reproducible, we include comprehensive information which involves disclosing code, datasets (we work with publicly accessible datasets and adhere to the ethical guidelines established by the datasets' creators), and other pertinent materials. The dataset in this study is designed for research on multimodal table question answering. It should be strictly used for research purposes, not for other applications As a result, the scientific community can verify and build upon our findings. The assertions made in this paper align with the outcomes of our experiments. But because black-box big language models are inherently stochastic, we have reduced variability by keeping the temperature constant. In order to ensure the reproducibility of our work, we provide comprehensive details about the prompting techniques utilized, models used, dataset splits, and annotations made.

\section*{Acknowledgement}
Research was sponsored by the Army Research Office and was accomplished under Grant Number W911NF-20-1-0080. The views and conclusions contained in this document are those of the authors and should not be interpreted as representing the official policies, either expressed or implied,
of the Army Research Office or the U.S. Government. The U.S. Government is authorized to reproduce and distribute reprints for Government purposes notwithstanding any copyright notation herein. This work was partially funded by ONR Contract N00014-19-1-2620. Lastly, we extend our appreciation to the reviewing team
for their insightful comments.

\bibliography{custom}

\begin{thebibliography}{38}
\providecommand{\natexlab}[1]{#1}

\bibitem[{Achiam et~al.(2023)Achiam, Adler, Agarwal, Ahmad, Akkaya, Aleman, Almeida, Altenschmidt, Altman, Anadkat et~al.}]{achiam2023gpt}
Josh Achiam, Steven Adler, Sandhini Agarwal, Lama Ahmad, Ilge Akkaya, Florencia~Leoni Aleman, Diogo Almeida, Janko Altenschmidt, Sam Altman, Shyamal Anadkat, et~al. 2023.
\newblock Gpt-4 technical report.
\newblock \emph{arXiv preprint arXiv:2303.08774}.

\bibitem[{Aly et~al.(2021)Aly, Guo, Schlichtkrull, Thorne, Vlachos, Christodoulopoulos, Cocarascu, and Mittal}]{aly-etal-2021-fact}
Rami Aly, Zhijiang Guo, Michael~Sejr Schlichtkrull, James Thorne, Andreas Vlachos, Christos Christodoulopoulos, Oana Cocarascu, and Arpit Mittal. 2021.
\newblock \href {https://doi.org/10.18653/v1/2021.fever-1.1} {The fact extraction and {VER}ification over unstructured and structured information ({FEVEROUS}) shared task}.
\newblock In \emph{Proceedings of the Fourth Workshop on Fact Extraction and VERification (FEVER)}, pages 1--13, Dominican Republic. Association for Computational Linguistics.

\bibitem[{Bai et~al.(2023)Bai, Bai, Yang, Wang, Tan, Wang, Lin, Zhou, and Zhou}]{bai2023qwen}
Jinze Bai, Shuai Bai, Shusheng Yang, Shijie Wang, Sinan Tan, Peng Wang, Junyang Lin, Chang Zhou, and Jingren Zhou. 2023.
\newblock Qwen-vl: A frontier large vision-language model with versatile abilities.
\newblock \emph{arXiv preprint arXiv:2308.12966}.

\bibitem[{Brown et~al.(2020)Brown, Mann, Ryder, Subbiah, Kaplan, Dhariwal, Neelakantan, Shyam, Sastry, Askell, Agarwal, Herbert-Voss, Krueger, Henighan, Child, Ramesh, Ziegler, Wu, Winter, Hesse, Chen, Sigler, Litwin, Gray, Chess, Clark, Berner, McCandlish, Radford, Sutskever, and Amodei}]{brown2020language}
Tom~B. Brown, Benjamin Mann, Nick Ryder, Melanie Subbiah, Jared Kaplan, Prafulla Dhariwal, Arvind Neelakantan, Pranav Shyam, Girish Sastry, Amanda Askell, Sandhini Agarwal, Ariel Herbert-Voss, Gretchen Krueger, Tom Henighan, Rewon Child, Aditya Ramesh, Daniel~M. Ziegler, Jeffrey Wu, Clemens Winter, Christopher Hesse, Mark Chen, Eric Sigler, Mateusz Litwin, Scott Gray, Benjamin Chess, Jack Clark, Christopher Berner, Sam McCandlish, Alec Radford, Ilya Sutskever, and Dario Amodei. 2020.
\newblock \href {https://arxiv.org/abs/2005.14165} {Language models are few-shot learners}.
\newblock \emph{Preprint}, arXiv:2005.14165.

\bibitem[{Chen et~al.(2020)Chen, Zha, Chen, Xiong, Wang, and Wang}]{chen-etal-2020-hybridqa}
Wenhu Chen, Hanwen Zha, Zhiyu Chen, Wenhan Xiong, Hong Wang, and William~Yang Wang. 2020.
\newblock \href {https://doi.org/10.18653/v1/2020.findings-emnlp.91} {{H}ybrid{QA}: A dataset of multi-hop question answering over tabular and textual data}.
\newblock In \emph{Findings of the Association for Computational Linguistics: EMNLP 2020}, pages 1026--1036, Online. Association for Computational Linguistics.

\bibitem[{Chen et~al.(2023)Chen, Hu, Luan, Sun, Changpinyo, Ritter, and Chang}]{Chen2023CanPV}
Yang Chen, Hexiang Hu, Yi~Luan, Haitian Sun, Soravit Changpinyo, Alan Ritter, and Ming-Wei Chang. 2023.
\newblock \href {https://api.semanticscholar.org/CorpusID:257102433} {Can pre-trained vision and language models answer visual information-seeking questions?}
\newblock \emph{ArXiv}, abs/2302.11713.

\bibitem[{Chen et~al.(2024)Chen, Wang, Tian, Ye, Gao, Cui, Tong, Hu, Luo, Ma et~al.}]{chen2024far}
Zhe Chen, Weiyun Wang, Hao Tian, Shenglong Ye, Zhangwei Gao, Erfei Cui, Wenwen Tong, Kongzhi Hu, Jiapeng Luo, Zheng Ma, et~al. 2024.
\newblock How far are we to gpt-4v? closing the gap to commercial multimodal models with open-source suites.
\newblock \emph{arXiv preprint arXiv:2404.16821}.

\bibitem[{Chen et~al.(2021)Chen, Chen, Smiley, Shah, Borova, Langdon, Moussa, Beane, Huang, Routledge, and Wang}]{chen-etal-2021-finqa}
Zhiyu Chen, Wenhu Chen, Charese Smiley, Sameena Shah, Iana Borova, Dylan Langdon, Reema Moussa, Matt Beane, Ting-Hao Huang, Bryan Routledge, and William~Yang Wang. 2021.
\newblock \href {https://doi.org/10.18653/v1/2021.emnlp-main.300} {{F}in{QA}: A dataset of numerical reasoning over financial data}.
\newblock In \emph{Proceedings of the 2021 Conference on Empirical Methods in Natural Language Processing}, pages 3697--3711, Online and Punta Cana, Dominican Republic. Association for Computational Linguistics.

\bibitem[{de~Faria et~al.(2023)de~Faria, Bastos, da~Silva, Fabris, Uchoa, Neto, and Santos}]{de2023visual}
Ana Cl{\'a}udia Akemi~Matsuki de~Faria, Felype de~Castro Bastos, Jos{\'e} Victor Nogueira~Alves da~Silva, Vitor~Lopes Fabris, Valeska de~Sousa Uchoa, D{\'e}cio Gon{\c{c}}alves de~Aguiar Neto, and Claudio Filipi Goncalves~dos Santos. 2023.
\newblock Visual question answering: A survey on techniques and common trends in recent literature.
\newblock \emph{arXiv preprint arXiv:2305.11033}.

\bibitem[{Fang et~al.(2024)Fang, Xu, Tan, Zhang, Hu, Qi, Nickleach, Socolinsky, Sengamedu, and Faloutsos}]{fang2024large}
Xi~Fang, Weijie Xu, Fiona~Anting Tan, Jiani Zhang, Ziqing Hu, Yanjun Qi, Scott Nickleach, Diego Socolinsky, Srinivasan Sengamedu, and Christos Faloutsos. 2024.
\newblock \href {https://arxiv.org/abs/2402.17944} {Large language models(llms) on tabular data: Prediction, generation, and understanding -- a survey}.
\newblock \emph{Preprint}, arXiv:2402.17944.

\bibitem[{Hong et~al.(2023)Hong, Wang, Lv, Xu, Yu, Ji, Wang, Wang, Dong, Ding et~al.}]{hong2023cogagent}
Wenyi Hong, Weihan Wang, Qingsong Lv, Jiazheng Xu, Wenmeng Yu, Junhui Ji, Yan Wang, Zihan Wang, Yuxiao Dong, Ming Ding, et~al. 2023.
\newblock Cogagent: A visual language model for gui agents.
\newblock \emph{arXiv preprint arXiv:2312.08914}.

\bibitem[{Jiang et~al.(2024{\natexlab{a}})Jiang, Sablayrolles, Roux, Mensch, Savary, Bamford, Chaplot, de~las Casas, Hanna, Bressand, Lengyel, Bour, Lample, Lavaud, Saulnier, Lachaux, Stock, Subramanian, Yang, Antoniak, Scao, Gervet, Lavril, Wang, Lacroix, and Sayed}]{jiang2024mixtral}
Albert~Q. Jiang, Alexandre Sablayrolles, Antoine Roux, Arthur Mensch, Blanche Savary, Chris Bamford, Devendra~Singh Chaplot, Diego de~las Casas, Emma~Bou Hanna, Florian Bressand, Gianna Lengyel, Guillaume Bour, Guillaume Lample, Lélio~Renard Lavaud, Lucile Saulnier, Marie-Anne Lachaux, Pierre Stock, Sandeep Subramanian, Sophia Yang, Szymon Antoniak, Teven~Le Scao, Théophile Gervet, Thibaut Lavril, Thomas Wang, Timothée Lacroix, and William~El Sayed. 2024{\natexlab{a}}.
\newblock \href {https://arxiv.org/abs/2401.04088} {Mixtral of experts}.
\newblock \emph{Preprint}, arXiv:2401.04088.

\bibitem[{Jiang et~al.(2024{\natexlab{b}})Jiang, He, Zeng, Wei, Ku, Liu, and Chen}]{jiang2024mantis}
Dongfu Jiang, Xuan He, Huaye Zeng, Cong Wei, Max Ku, Qian Liu, and Wenhu Chen. 2024{\natexlab{b}}.
\newblock \href {https://arxiv.org/abs/2405.01483} {Mantis: Interleaved multi-image instruction tuning}.
\newblock \emph{Preprint}, arXiv:2405.01483.

\bibitem[{Jin et~al.(2022)Jin, Siebert, Li, and Chen}]{table_survey}
Nengzheng Jin, Joanna Siebert, Dongfang Li, and Qingcai Chen. 2022.
\newblock A survey on table question answering: Recent advances.
\newblock In \emph{Knowledge Graph and Semantic Computing: Knowledge Graph Empowers the Digital Economy}, pages 174--186, Singapore. Springer Nature Singapore.

\bibitem[{Kim et~al.(2024)Kim, Yim, and Song}]{kim2024tablevqa}
Yoonsik Kim, Moonbin Yim, and Ka~Yeon Song. 2024.
\newblock Tablevqa-bench: A visual question answering benchmark on multiple table domains.
\newblock \emph{arXiv preprint arXiv:2404.19205}.

\bibitem[{Lerner et~al.(2022)Lerner, Ferret, Guinaudeau, Le~Borgne, Besan{\c{c}}on, Moreno, and Lov{\'o}n~Melgarejo}]{lerner2022viquae}
Paul Lerner, Olivier Ferret, Camille Guinaudeau, Herv{\'e} Le~Borgne, Romaric Besan{\c{c}}on, Jos{\'e}~G Moreno, and Jes{\'u}s Lov{\'o}n~Melgarejo. 2022.
\newblock Viquae, a dataset for knowledge-based visual question answering about named entities.
\newblock In \emph{Proceedings of the 45th International ACM SIGIR Conference on Research and Development in Information Retrieval}, pages 3108--3120.

\bibitem[{Li et~al.(2022)Li, Li, and Nie}]{li-etal-2022-mmcoqa}
Yongqi Li, Wenjie Li, and Liqiang Nie. 2022.
\newblock \href {https://doi.org/10.18653/v1/2022.acl-long.290} {{MMC}o{QA}: Conversational question answering over text, tables, and images}.
\newblock In \emph{Proceedings of the 60th Annual Meeting of the Association for Computational Linguistics (Volume 1: Long Papers)}, pages 4220--4231, Dublin, Ireland. Association for Computational Linguistics.

\bibitem[{Lin(2004)}]{lin-2004-rouge}
Chin-Yew Lin. 2004.
\newblock \href {https://aclanthology.org/W04-1013} {{ROUGE}: A package for automatic evaluation of summaries}.
\newblock In \emph{Text Summarization Branches Out}, pages 74--81, Barcelona, Spain. Association for Computational Linguistics.

\bibitem[{Lu et~al.(2023)Lu, Qiu, Chang, Wu, Zhu, Rajpurohit, Clark, and Kalyan}]{lu2023dynamic}
Pan Lu, Liang Qiu, Kai-Wei Chang, Ying~Nian Wu, Song-Chun Zhu, Tanmay Rajpurohit, Peter Clark, and Ashwin Kalyan. 2023.
\newblock Dynamic prompt learning via policy gradient for semi-structured mathematical reasoning.
\newblock In \emph{International Conference on Learning Representations (ICLR)}.

\bibitem[{Marino et~al.(2019)Marino, Rastegari, Farhadi, and Mottaghi}]{okvqa}
Kenneth Marino, Mohammad Rastegari, Ali Farhadi, and Roozbeh Mottaghi. 2019.
\newblock \href {https://arxiv.org/abs/1906.00067} {Ok-vqa: A visual question answering benchmark requiring external knowledge}.
\newblock \emph{Preprint}, arXiv:1906.00067.

\bibitem[{Mensink et~al.(2023)Mensink, Uijlings, Castrejon, Goel, Cadar, Zhou, Sha, Araujo, and Ferrari}]{encyclopedic_vqa}
T.~Mensink, J.~Uijlings, L.~Castrejon, A.~Goel, F.~Cadar, H.~Zhou, F.~Sha, A.~Araujo, and V.~Ferrari. 2023.
\newblock \href {https://doi.org/10.1109/ICCV51070.2023.00289} {Encyclopedic vqa: Visual questions about detailed properties of fine-grained categories}.
\newblock In \emph{2023 IEEE/CVF International Conference on Computer Vision (ICCV)}, pages 3090--3101, Los Alamitos, CA, USA. IEEE Computer Society.

\bibitem[{Moon et~al.(2018)Moon, Neves, and Carvalho}]{moon-etal-2018-multimodal-named}
Seungwhan Moon, Leonardo Neves, and Vitor Carvalho. 2018.
\newblock \href {https://doi.org/10.18653/v1/P18-1186} {Multimodal named entity disambiguation for noisy social media posts}.
\newblock In \emph{Proceedings of the 56th Annual Meeting of the Association for Computational Linguistics (Volume 1: Long Papers)}, pages 2000--2008, Melbourne, Australia. Association for Computational Linguistics.

\bibitem[{M{\"u}ller et~al.(2021)M{\"u}ller, Eisenschlos, and Krichene}]{muller-etal-2021-tapas}
Thomas M{\"u}ller, Julian Eisenschlos, and Syrine Krichene. 2021.
\newblock \href {https://doi.org/10.18653/v1/2021.semeval-1.51} {{TAPAS} at {S}em{E}val-2021 task 9: Reasoning over tables with intermediate pre-training}.
\newblock In \emph{Proceedings of the 15th International Workshop on Semantic Evaluation (SemEval-2021)}, pages 423--430, Online. Association for Computational Linguistics.

\bibitem[{Nan et~al.(2022)Nan, Hsieh, Mao, Lin, Verma, Zhang, Kry{\'s}ci{\'n}ski, Schoelkopf, Kong, Tang, Mutuma, Rosand, Trindade, Bandaru, Cunningham, Xiong, Radev, and Radev}]{nan-etal-2022-fetaqa}
Linyong Nan, Chiachun Hsieh, Ziming Mao, Xi~Victoria Lin, Neha Verma, Rui Zhang, Wojciech Kry{\'s}ci{\'n}ski, Hailey Schoelkopf, Riley Kong, Xiangru Tang, Mutethia Mutuma, Ben Rosand, Isabel Trindade, Renusree Bandaru, Jacob Cunningham, Caiming Xiong, Dragomir Radev, and Dragomir Radev. 2022.
\newblock \href {https://doi.org/10.1162/tacl_a_00446} {{F}e{T}a{QA}: Free-form table question answering}.
\newblock \emph{Transactions of the Association for Computational Linguistics}, 10:35--49.

\bibitem[{Pasupat and Liang(2015)}]{pasupat-liang-2015-compositional}
Panupong Pasupat and Percy Liang. 2015.
\newblock \href {https://doi.org/10.3115/v1/P15-1142} {Compositional semantic parsing on semi-structured tables}.
\newblock In \emph{Proceedings of the 53rd Annual Meeting of the Association for Computational Linguistics and the 7th International Joint Conference on Natural Language Processing (Volume 1: Long Papers)}, pages 1470--1480, Beijing, China. Association for Computational Linguistics.

\bibitem[{Penamakuri et~al.(2023)Penamakuri, Gupta, Gupta, and Mishra}]{10.24963/ijcai.2023/146}
Abhirama~Subramanyam Penamakuri, Manish Gupta, Mithun~Das Gupta, and Anand Mishra. 2023.
\newblock \href {https://doi.org/10.24963/ijcai.2023/146} {Answer mining from a pool of images: towards retrieval-based visual question answering}.
\newblock In \emph{Proceedings of the Thirty-Second International Joint Conference on Artificial Intelligence}, IJCAI '23.

\bibitem[{Schwenk et~al.(2022)Schwenk, Khandelwal, Clark, Marino, and Mottaghi}]{aokvqa}
Dustin Schwenk, Apoorv Khandelwal, Christopher Clark, Kenneth Marino, and Roozbeh Mottaghi. 2022.
\newblock \href {https://api.semanticscholar.org/CorpusID:249375629} {A-okvqa: A benchmark for visual question answering using world knowledge}.
\newblock In \emph{European Conference on Computer Vision}.

\bibitem[{Talmor et~al.(2021)Talmor, Yoran, Catav, Lahav, Wang, Asai, Ilharco, Hajishirzi, and Berant}]{Talmor2021MultiModalQACQ}
Alon Talmor, Ori Yoran, Amnon Catav, Dan Lahav, Yizhong Wang, Akari Asai, Gabriel Ilharco, Hannaneh Hajishirzi, and Jonathan Berant. 2021.
\newblock \href {https://api.semanticscholar.org/CorpusID:233219849} {Multimodalqa: Complex question answering over text, tables and images}.
\newblock \emph{ArXiv}, abs/2104.06039.

\bibitem[{Team et~al.(2023)Team, Anil, Borgeaud, Wu, Alayrac, Yu, Soricut, Schalkwyk, Dai, Hauth et~al.}]{team2023gemini}
Gemini Team, Rohan Anil, Sebastian Borgeaud, Yonghui Wu, Jean-Baptiste Alayrac, Jiahui Yu, Radu Soricut, Johan Schalkwyk, Andrew~M Dai, Anja Hauth, et~al. 2023.
\newblock Gemini: a family of highly capable multimodal models.
\newblock \emph{arXiv preprint arXiv:2312.11805}.

\bibitem[{Tian et~al.(2024)Tian, Zhu, Xiong, Wang, Chen, Wang, Chen, Lu, Lu, Zhou, Li, Qiao, and Dai}]{tian2024mminterleaved}
Changyao Tian, Xizhou Zhu, Yuwen Xiong, Weiyun Wang, Zhe Chen, Wenhai Wang, Yuntao Chen, Lewei Lu, Tong Lu, Jie Zhou, Hongsheng Li, Yu~Qiao, and Jifeng Dai. 2024.
\newblock \href {https://arxiv.org/abs/2401.10208} {Mm-interleaved: Interleaved image-text generative modeling via multi-modal feature synchronizer}.
\newblock \emph{Preprint}, arXiv:2401.10208.

\bibitem[{Touvron et~al.(2023)Touvron, Lavril, Izacard, Martinet, Lachaux, Lacroix, Rozi{\`e}re, Goyal, Hambro, Azhar et~al.}]{touvron2023llama}
Hugo Touvron, Thibaut Lavril, Gautier Izacard, Xavier Martinet, Marie-Anne Lachaux, Timoth{\'e}e Lacroix, Baptiste Rozi{\`e}re, Naman Goyal, Eric Hambro, Faisal Azhar, et~al. 2023.
\newblock Llama: Open and efficient foundation language models.
\newblock \emph{arXiv preprint arXiv:2302.13971}.

\bibitem[{Wang et~al.(2021)Wang, Mahajan, Danilevsky, and Rosenthal}]{wang-etal-2021-semeval}
Nancy X.~R. Wang, Diwakar Mahajan, Marina Danilevsky, and Sara Rosenthal. 2021.
\newblock \href {https://doi.org/10.18653/v1/2021.semeval-1.39} {{S}em{E}val-2021 task 9: Fact verification and evidence finding for tabular data in scientific documents ({SEM}-{TAB}-{FACTS})}.
\newblock In \emph{Proceedings of the 15th International Workshop on Semantic Evaluation (SemEval-2021)}, pages 317--326, Online. Association for Computational Linguistics.

\bibitem[{Wang et~al.(2022)Wang, Tian, Gui, Li, Wang, Yan, Chen, and Xiao}]{wang-etal-2022-wikidiverse}
Xuwu Wang, Junfeng Tian, Min Gui, Zhixu Li, Rui Wang, Ming Yan, Lihan Chen, and Yanghua Xiao. 2022.
\newblock \href {https://doi.org/10.18653/v1/2022.acl-long.328} {{W}iki{D}iverse: A multimodal entity linking dataset with diversified contextual topics and entity types}.
\newblock In \emph{Proceedings of the 60th Annual Meeting of the Association for Computational Linguistics (Volume 1: Long Papers)}, pages 4785--4797, Dublin, Ireland. Association for Computational Linguistics.

\bibitem[{Wei et~al.(2023)Wei, Wang, Schuurmans, Bosma, Ichter, Xia, Chi, Le, and Zhou}]{wei2023chainofthought}
Jason Wei, Xuezhi Wang, Dale Schuurmans, Maarten Bosma, Brian Ichter, Fei Xia, Ed~Chi, Quoc Le, and Denny Zhou. 2023.
\newblock \href {https://arxiv.org/abs/2201.11903} {Chain-of-thought prompting elicits reasoning in large language models}.
\newblock \emph{Preprint}, arXiv:2201.11903.

\bibitem[{Zhao et~al.(2022)Zhao, Li, Li, and Zhang}]{zhao-etal-2022-multihiertt}
Yilun Zhao, Yunxiang Li, Chenying Li, and Rui Zhang. 2022.
\newblock \href {https://doi.org/10.18653/v1/2022.acl-long.454} {{M}ulti{H}iertt: Numerical reasoning over multi hierarchical tabular and textual data}.
\newblock In \emph{Proceedings of the 60th Annual Meeting of the Association for Computational Linguistics (Volume 1: Long Papers)}, pages 6588--6600, Dublin, Ireland. Association for Computational Linguistics.

\bibitem[{Zheng et~al.(2024)Zheng, Feng, Si, She, Lin, Jiang, and Wang}]{zheng2024multimodal}
Mingyu Zheng, Xinwei Feng, Qingyi Si, Qiaoqiao She, Zheng Lin, Wenbin Jiang, and Weiping Wang. 2024.
\newblock \href {https://arxiv.org/abs/2406.08100} {Multimodal table understanding}.
\newblock \emph{Preprint}, arXiv:2406.08100.

\bibitem[{Zhong et~al.(2017)Zhong, Xiong, and Socher}]{zhong2017seq2sql}
Victor Zhong, Caiming Xiong, and Richard Socher. 2017.
\newblock Seq2sql: Generating structured queries from natural language using reinforcement learning.
\newblock \emph{arXiv preprint arXiv:1709.00103}.

\bibitem[{Zhu et~al.(2021)Zhu, Lei, Huang, Wang, Zhang, Lv, Feng, and Chua}]{zhu-etal-2021-tat}
Fengbin Zhu, Wenqiang Lei, Youcheng Huang, Chao Wang, Shuo Zhang, Jiancheng Lv, Fuli Feng, and Tat-Seng Chua. 2021.
\newblock \href {https://doi.org/10.18653/v1/2021.acl-long.254} {{TAT}-{QA}: A question answering benchmark on a hybrid of tabular and textual content in finance}.
\newblock In \emph{Proceedings of the 59th Annual Meeting of the Association for Computational Linguistics and the 11th International Joint Conference on Natural Language Processing (Volume 1: Long Papers)}, pages 3277--3287, Online. Association for Computational Linguistics.

\end{thebibliography}

% \newpage \pagebreak \cleardoublepage

\appendix
\section{Dataset Examples}
\begin{figure}[htbp]
    \centering
    \includegraphics[width=0.47\textwidth]{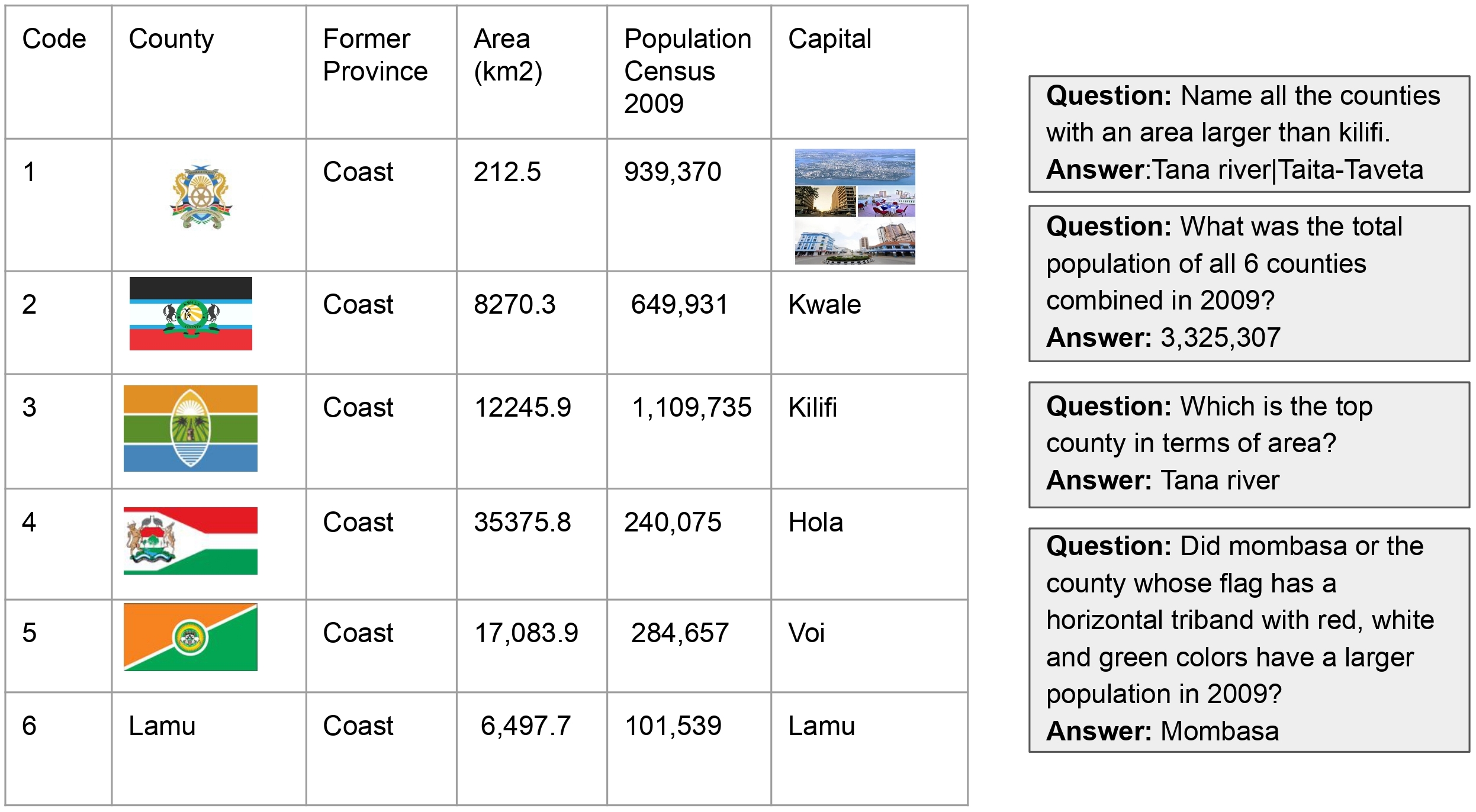}
    \caption{WikiTableQuestions Dataset Example}
    \label{fig:WTQ-ex}
\end{figure}

\begin{figure}[htbp]
    \centering
    \includegraphics[width=0.47\textwidth]{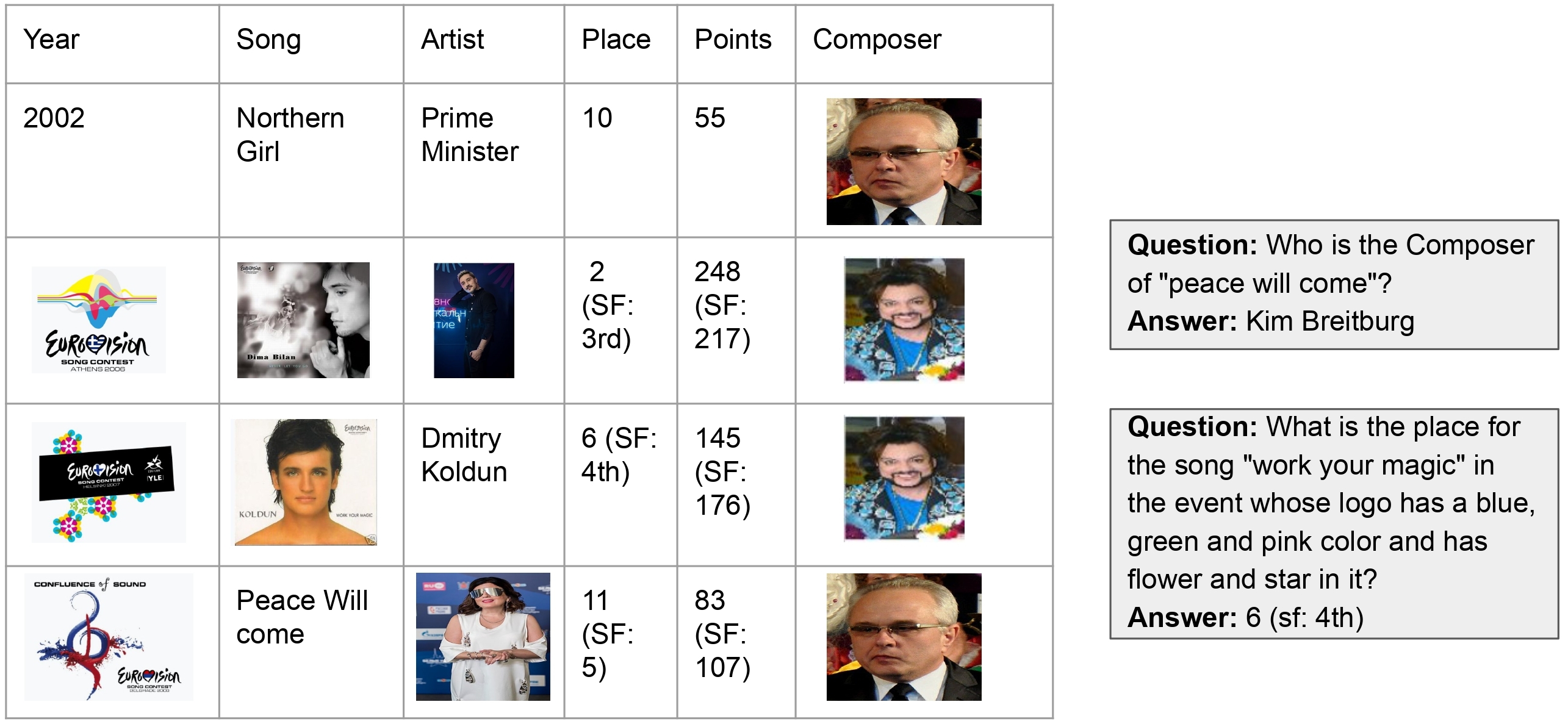}
    \caption{WikiSQL Dataset Example}
    \label{fig:WSL-ex}
\end{figure}

\begin{figure}[htbp]
    \centering
    \includegraphics[width=0.47\textwidth]{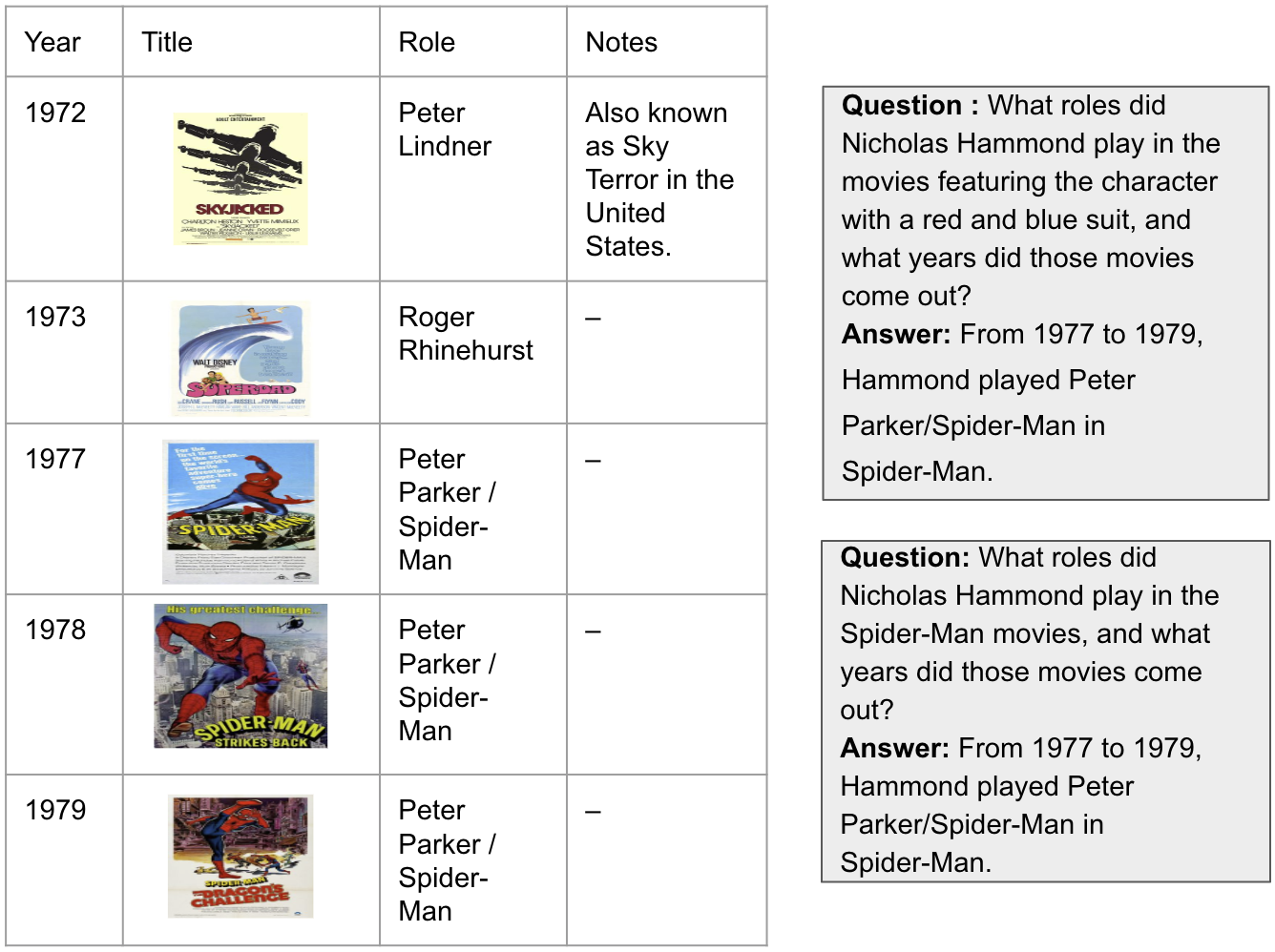}
    \caption{FeTaQA Dataset Example}
    \label{fig:FTQ-ex}
\end{figure}

\begin{figure}[htbp]
    \centering
    \includegraphics[width=0.47\textwidth]{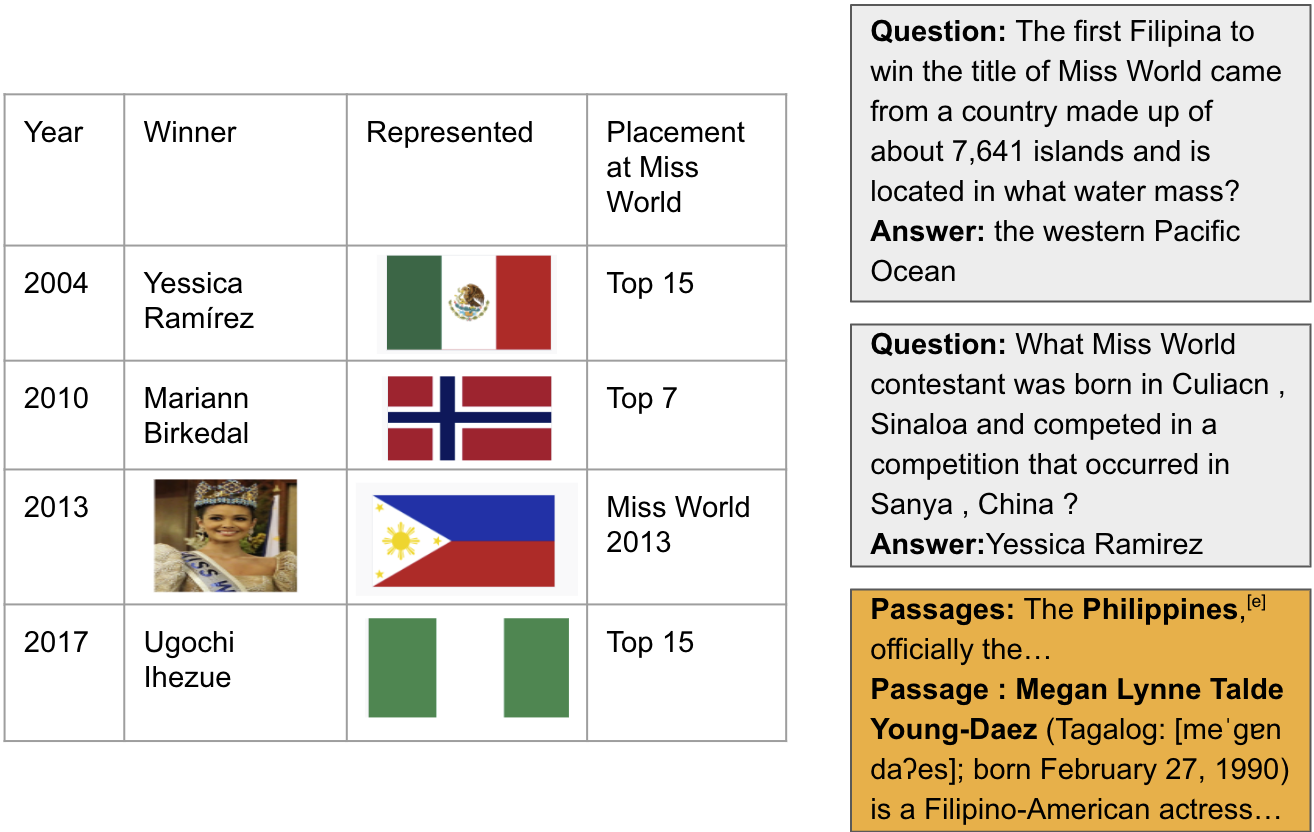}
    \caption{HybridQA Dataset Example}
    \label{fig:HQA-ex}
\end{figure}

To demonstrate the quality and features of our created {\sc MMTabQA} we provide table examples along with question - answer pair from all the four datasets. Fig. \ref{fig:WTQ-ex} shows the examples from WikiTableQuestions dataset, Fig. \ref{fig:WSL-ex} shows the examples from WikiSQL dataset, Fig. \ref{fig:FTQ-ex} shows the examples from FeTaQA dataset, Fig. \ref{fig:HQA-ex} shows the examples from HybridQA dataset.

\section{HybridQA Experiments}
% Please add the following required packages to your document preamble:
% \usepackage{booktabs}
% \usepackage{graphicx}
% \usepackage[table,xcdraw]{xcolor}
% Beamer presentation requires \usepackage{colortbl} instead of \usepackage[table,xcdraw]{xcolor}
\begin{table*}[ht]
\small
\setlength{\tabcolsep}{3.5pt}
\resizebox{\textwidth}{!}{%
\begin{tabular}{@{}lllllllllllll@{}}
\toprule
\multicolumn{1}{l|}{Models} &
  \multicolumn{3}{c|}{EQ} &
  \multicolumn{3}{c|}{AQ} &
  \multicolumn{3}{c|}{IQ} &
  \multicolumn{3}{c}{VQ} \\ \midrule
\multicolumn{1}{l|}{Metrics} &
  \multicolumn{1}{l|}{EM} &
  \multicolumn{1}{l|}{SSM} &
  \multicolumn{1}{l|}{F1} &
  \multicolumn{1}{l|}{EM} &
  \multicolumn{1}{l|}{SSM} &
  \multicolumn{1}{l|}{F1} &
  \multicolumn{1}{l|}{EM} &
  \multicolumn{1}{l|}{SSM} &
  \multicolumn{1}{l|}{F1} &
  \multicolumn{1}{l|}{EM} &
  \multicolumn{1}{l|}{SSM} &
  F1 \\ \midrule
\multicolumn{13}{c}{Oracle-Entity Replaced Baseline} \\ \midrule
\multicolumn{1}{l|}{Gemini 1.5 Flash} &
  \cellcolor[HTML]{FFCCC9}63.71 &
  \cellcolor[HTML]{FFCCC9}77.14 &
  \multicolumn{1}{l|}{\cellcolor[HTML]{FFCCC9}0.77} &
  46.49 &
  53.65 &
  \multicolumn{1}{l|}{0.54} &
  60.83 &
  71.59 &
  \multicolumn{1}{l|}{0.73} &
  - &
  - &
  - \\
\multicolumn{1}{l|}{LLAMA 3 70B} &
  61.76 &
  75.47 &
  \multicolumn{1}{l|}{0.74} &
  \cellcolor[HTML]{FFCCC9}55.25 &
  \cellcolor[HTML]{FFCCC9}{\color[HTML]{333333} 61.58} &
  \multicolumn{1}{l|}{\cellcolor[HTML]{FFCCC9}{\color[HTML]{333333} 0.62}} &
  \cellcolor[HTML]{FFCCC9}{\color[HTML]{333333} 60.86} &
  \cellcolor[HTML]{FFCCC9}{\color[HTML]{333333} 73.09} &
  \multicolumn{1}{l|}{\cellcolor[HTML]{FFCCC9}{\color[HTML]{333333} 0.72}} &
  - &
  - &
  - \\
\multicolumn{1}{l|}{Mixtral 8x7B} &
  46.00 &
  63.90 &
  \multicolumn{1}{l|}{0.63} &
  34.97 &
  53.90 &
  \multicolumn{1}{l|}{0.45} &
  33.67 &
  67.63 &
  \multicolumn{1}{l|}{0.50} &
  - &
  - &
  - \\ \midrule
\multicolumn{13}{c}{Partial Input Baseline} \\ \midrule
\multicolumn{1}{l|}{Gemini 1.5 Flash} &
  \cellcolor[HTML]{FFCCC9}59.71 &
  \cellcolor[HTML]{FFCCC9}71.57 &
  \multicolumn{1}{l|}{\cellcolor[HTML]{FFCCC9}0.71} &
  28.71 &
  32.43 &
  \multicolumn{1}{l|}{0.34} &
  \cellcolor[HTML]{FFCCC9}59.14 &
  \cellcolor[HTML]{FFCCC9}69.71 &
  \multicolumn{1}{l|}{\cellcolor[HTML]{FFCCC9}0.70} &
  43.29 &
  53.46 &
  0.54 \\
\multicolumn{1}{l|}{LLAMA 3 70B} &
  55.56 &
  70.13 &
  \multicolumn{1}{l|}{0.68} &
  \cellcolor[HTML]{FFCCC9}34.01 &
  \cellcolor[HTML]{FFCCC9}39.91 &
  \multicolumn{1}{l|}{\cellcolor[HTML]{FFCCC9}0.41} &
  56.83 &
  68.63 &
  \multicolumn{1}{l|}{0.69} &
  \cellcolor[HTML]{FFCCC9}45.04 &
  \cellcolor[HTML]{FFCCC9}59.92 &
  \cellcolor[HTML]{FFCCC9}0.57 \\
\multicolumn{1}{l|}{Mixtral 8x7B} &
  40.23 &
  58.16 &
  \multicolumn{1}{l|}{0.56} &
  26.01 &
  30.92 &
  \multicolumn{1}{l|}{0.33} &
  45.44 &
  59.04 &
  \multicolumn{1}{l|}{0.58} &
  28.90 &
  41.37 &
  0.42 \\ \midrule
  
  \multicolumn{13}{c}{Image-captioning Baseline} \\ \midrule
\multicolumn{1}{l|}{Gemini-1.5 Flash} &
  59.57 &
  71.85 &
  \multicolumn{1}{l|}{0.71} &
  39.57&
  43.00 &
  \multicolumn{1}{l|}{0.46} &
  59.14 &
  67.28 &
  \multicolumn{1}{l|}{0.69} &
  44.91 &
  52.64 &
  0.55 \\ \midrule

\multicolumn{13}{c}{Table-as-an-Image} \\ \midrule
\multicolumn{1}{l|}{Gemini 1.5 flash} &
  48.50 &
  67.67 &
  \multicolumn{1}{l|}{0.64} &
  26.14 &
  33.28 &
  \multicolumn{1}{l|}{0.34} &
  47.78 &
  68.10 &
  \multicolumn{1}{l|}{0.63} &
  42.07 &
  58.54 &
  0.56 \\
\multicolumn{1}{l|}{GPT-4o} &
  \cellcolor[HTML]{FFCCC9}62.05 &
  \cellcolor[HTML]{FFCCC9}76.31 &
  \multicolumn{1}{l|}{\cellcolor[HTML]{FFCCC9}0.75} &
  \cellcolor[HTML]{FFCCC9}48.00 &
  \cellcolor[HTML]{FFCCC9}53.00 &
  \multicolumn{1}{l|}{\cellcolor[HTML]{FFCCC9}0.56} &
  \cellcolor[HTML]{FFCCC9}64.46 &
  \cellcolor[HTML]{FFCCC9}75.50 &
  \multicolumn{1}{l|}{\cellcolor[HTML]{FFCCC9}0.76} &
  \cellcolor[HTML]{FFCCC9}50.81 &
  \cellcolor[HTML]{FFCCC9}61.18 &
  \cellcolor[HTML]{FFCCC9}0.63 \\
\multicolumn{1}{l|}{Qwen-VL-chat} &
  12.81 &
  16.08 &
  \multicolumn{1}{l|}{0.17} &
  7.31 &
  10.32 &
  \multicolumn{1}{l|}{0.13} &
  9.87 &
  13.16 &
  \multicolumn{1}{l|}{0.14} &
  6.72 &
  11.00 &
  0.11 \\
\multicolumn{1}{l|}{CogAgent-VQA} &
  15.74 &
  19.64 &
  \multicolumn{1}{l|}{0.211} &
   9.58 &
  11.79 &
  \multicolumn{1}{l|}{0.15} &
    15.29 &
   20.00&
  \multicolumn{1}{l|}{0.21} &
  11.39 &
  14.55 &
  0.17 \\
\multicolumn{1}{l|}{Intern-VLM-4khd} &
  43.60 &
  59.74 &
  \multicolumn{1}{l|}{0.57} &
  22.25 &
  26.59 &
  \multicolumn{1}{l|}{0.29} &
  43.95 &
  57.10 &
  \multicolumn{1}{l|}{0.56} &
  35.48 &
  44.81 &
  0.46 \\ \midrule
\multicolumn{13}{c}{Interleaved Text-Image Baseline} \\ \midrule
\multicolumn{1}{l|}{Gemini 1.5 Flash} &
  45.94 &
  71.21 &
  \multicolumn{1}{l|}{0.63} &
  25.77 &
  36.79 &
  \multicolumn{1}{l|}{0.35} &
  49.13 &
  \cellcolor[HTML]{FFCCC9}74.05 &
  \multicolumn{1}{l|}{0.66} &
  39.76 &
  60.00 &
  0.57 \\
\multicolumn{1}{l|}{GPT4o} &
  \cellcolor[HTML]{FFCCC9}59.80 &
  \cellcolor[HTML]{FFCCC9}80.20 &
  \multicolumn{1}{l|}{\cellcolor[HTML]{FFCCC9}0.76} &
  \cellcolor[HTML]{FFCCC9}43.40 &
  \cellcolor[HTML]{FFCCC9}49.80 &
  \multicolumn{1}{l|}{\cellcolor[HTML]{FFCCC9}0.52} &
  \cellcolor[HTML]{FFCCC9}51.80 &
  69.20 &
  \multicolumn{1}{l|}{\cellcolor[HTML]{FFCCC9}0.68} &
  \cellcolor[HTML]{FFCCC9}45.53 &
  \cellcolor[HTML]{FFCCC9}60.77 &
  \cellcolor[HTML]{FFCCC9}0.61 \\
\multicolumn{1}{l|}{Qwen-VL} &
  1.66 &
  5.80 &
  \multicolumn{1}{l|}{0.03} &
  2.12 &
  7.43 &
  \multicolumn{1}{l|}{0.05} &
  1.64 &
  10.27 &
  \multicolumn{1}{l|}{0.03} &
  1.03 &
  4.52 &
  0.02 \\
\multicolumn{1}{l|}{Idefics-Mantis} &
  0.58 &
  2.05 &
  \multicolumn{1}{l|}{0.02} &
  0.00 &
  0.92 &
  \multicolumn{1}{l|}{0.02} &
  1.11 &
  3.78 &
  \multicolumn{1}{l|}{0.02} &
  0.54 &
  1.34 &
  0.01
\end{tabular}
}
\caption{Detailed results on sampled subset of HybridQA. EM - Exact Match, SSM - Substring Match, F1 - F1 score. EQ - Explicit Questions, AQ - Answer-Mention Questions, IQ - Implicit Questions, VQ - Visual Questions. Best performing models are highlighted in red.}
%\vspace{-0.75em}
\label{tab:table-hybridqa}
\end{table*}

In addition to the data sources analyzed previously, we have incorporated the HybridQA dataset to enhance our proposed task with a question-answering component that requires reasoning over heterogeneous information. HybridQA aligns each question with a Wikipedia table and multiple free-form text corpora linked to the entities within the table. The design of the questions necessitates the aggregation of both tabular and textual information, rendering them unanswerable if either form is lacking. Oracle retrieval is employed to obtain the relevant passages for question-answering tasks.

We benchmark our augmented dataset utilizing the four approaches outlined earlier: text-only baselines (Partial Output Baseline and Oracle Entity Baseline), the Table-as-an-Image approach, and the Interleaved Text-Image approach, as presented in Table \ref{tab:table-hybridqa}. Exact Match, Substring Match, and F-1 Score are the metrics employed to evaluate the model's results. For the purpose of analysis, we will primarily focus on Substring Match.

Examining different approaches, Llama 3-70B and Gemini-1.5 Flash demonstrate comparable performance on text-only baseline models, indicating that the open-source model is equally capable as the closed-source model. Mistral 8x7B, however, underperforms, which can be attributed to the fewer parameters it contains.

In multimodal baselines, GPT-4o exhibits the best performance, with Gemini-1.5 Flash being a close second for both Table as Image and Interleaved Text-Image approaches. Open-source models display an interesting trend for these tasks. In the Table as Image approach, CogAgent-VQA and Intern-VLM-xcomposer-4khd provide decent performance, comparable to closed-source models, whereas Qwen-VL seems to underperform, likely due to the same parameter-related issues faced by Mistral in the text-only baseline.

A clear distinction emerges for the Interleaved Text-Image approach: closed-source models outperform open-source models, with GPT-4o being the best. Open-source models struggle to handle and infer from multiple images, and their smaller size further limits their performance.

A major observation is that for both text-only approaches, which represent the boundary values for the task, the performance metrics are quite close. Overall, the models demonstrate decent performance across all tasks and baselines. This can be attributed to the use of passages as additional context, which facilitates entity disambiguation for the models. However, this approach undermines the primary objective of our task, which is to challenge the models' ability to reason over heterogeneous information without relying heavily on supplementary textual context.

\section{Additional Metrics}
We present a detailed benchmark report for our dataset across various data sources. For WikiSQL (Fig. \ref{tab:WSL}) and WikiTableQuestions (Fig. \ref{tab:my-WTQ}), we report Exact Match (EM), Substring Match (SSM), and F1 Score. For FeTaQA (Fig. \ref{tab:table-fetaqa}), we include BLEU Score, ROUGE-1 (R-1), ROUGE-2 (R-2), and ROUGE-L (R-L). These detailed evaluations provide a thorough understanding of the models' capabilities and performance variations across different datasets, highlighting their strengths \& weakness.
% Please add the following required packages to your document preamble:
% \usepackage{graphicx}
% \usepackage[table,xcdraw]{xcolor}
% Beamer presentation requires \usepackage{colortbl} instead of \usepackage[table,xcdraw]{xcolor}

\begin{table*}[ht]
\small
\setlength{\tabcolsep}{3.5pt}
\resizebox{\textwidth}{!}{%
\begin{tabular}{@{}lllllllllllll@{}}
\toprule
\multicolumn{1}{l|}{Question-Type} &
  \multicolumn{3}{c|}{EQ} &
  \multicolumn{3}{c|}{IQ} &
  \multicolumn{3}{c|}{AQ} &
  \multicolumn{3}{c}{VQ}
   \\  \midrule
\multicolumn{1}{l|}{Metrics} &
  \multicolumn{1}{c|}{EM} &
  \multicolumn{1}{c|}{SSM} &
  \multicolumn{1}{c|}{F1} &
  \multicolumn{1}{c|}{EM} &
  \multicolumn{1}{c|}{SSM} &
  \multicolumn{1}{c|}{F1} &
  \multicolumn{1}{c|}{EM} &
  \multicolumn{1}{c|}{SSM} &
  \multicolumn{1}{c|}{F1} &
  \multicolumn{1}{c|}{EM} &
  \multicolumn{1}{c|}{SSM} &
  F1 \\ \midrule
\multicolumn{13}{c}{Partial Input Baseline} \\ \midrule
\multicolumn{1}{l|}{Gemini 1.5 Flash} &
  38.16 &
  40.99 &
  \multicolumn{1}{c|}{0.42} &
  \cellcolor[HTML]{FFCCC9}25.05 &
  \cellcolor[HTML]{FFCCC9}27.39 &
  \multicolumn{1}{c|}{\cellcolor[HTML]{FFCCC9}0.35} &
  \cellcolor[HTML]{FFCCC9}46.70 &
  \cellcolor[HTML]{FFCCC9}48.96 &
  \multicolumn{1}{c|}{\cellcolor[HTML]{FFCCC9}0.48} &
  28.40 &
  31.40 &
  \multicolumn{1}{c}{0.31}
   \\
\multicolumn{1}{l|}{LLAMA 3 70B} &
  \cellcolor[HTML]{FFCCC9}39.43 &
  \cellcolor[HTML]{FFCCC9}41.13 &
  \multicolumn{1}{c|}{\cellcolor[HTML]{FFCCC9}0.43} &
  24.50 &
  26.49 &
  \multicolumn{1}{c|}{0.32} &
  41.49 &
  43.75 &
  \multicolumn{1}{c|}{0.43} &
  \cellcolor[HTML]{FFCCC9}29.40 &
  \cellcolor[HTML]{FFCCC9}31.80 &
  \multicolumn{1}{c}{\cellcolor[HTML]{FFCCC9}0.34}
   \\
\multicolumn{1}{l|}{Mixtral 8x7B} &
  23.58 &
  26.56 &
  \multicolumn{1}{c|}{0.26} &
  8.83 &
  9.91 &
  \multicolumn{1}{c|}{0.13} &
  28.17 &
  30.26 &
  \multicolumn{1}{c|}{0.30} &
  17.60 &
  20.20 &
  \multicolumn{1}{c}{0.20}
   \\ \midrule
\multicolumn{13}{c}{Oracle-Entity Replaced Baseline} \\ \midrule
\multicolumn{1}{l|}{Gemini 1.5 Flash} &
  \cellcolor[HTML]{FFCCC9}71.21 &
  74.89 &
  \multicolumn{1}{c|}{\cellcolor[HTML]{FFCCC9}0.74} &
  \cellcolor[HTML]{FFCCC9}75.68 &
  \cellcolor[HTML]{FFCCC9}78.20 &
  \multicolumn{1}{c|}{\cellcolor[HTML]{FFCCC9}0.83} &
  \cellcolor[HTML]{FFCCC9}53.82 &
  54.86 &
  \multicolumn{1}{c|}{\cellcolor[HTML]{FFCCC9}0.55} &
  - &
  - &
  \multicolumn{1}{c}{-}
   \\
\multicolumn{1}{l|}{LLAMA 3.00 70B} &
  53.62 &
  \cellcolor[HTML]{FFCCC9}75.74 &
  \multicolumn{1}{c|}{0.61} &
  53.69 &
  75.32 &
  \multicolumn{1}{c|}{0.67} &
  45.49 &
  \cellcolor[HTML]{FFCCC9}58.85 &
  \multicolumn{1}{c|}{0.49} &
  - &
  - &
  \multicolumn{1}{c}{-}
   \\
\multicolumn{1}{l|}{Mixtral 8x7B} &
  48.79 &
  54.89 &
  \multicolumn{1}{c|}{0.54} &
  48.47 &
  53.87 &
  \multicolumn{1}{c|}{0.57} &
  37.57 &
  40.70 &
  \multicolumn{1}{c|}{0.40} &
  - &
  - &
  \multicolumn{1}{c}{-}
   \\ \midrule
\multicolumn{13}{c}{Image Captioning Baseline} \\ \midrule
\multicolumn{1}{l|}{Gemini 1.5 Flash} &
  48.65 &
  52.34 &
  \multicolumn{1}{c|}{0.52} &
  34.41 &
  42.16 &
  \multicolumn{1}{c|}{0.48} &
  48.96 &
  51.39 &
  \multicolumn{1}{c|}{0.50} &
  38.20 &
  42.20 &
  \multicolumn{1}{c}{0.43}
   \\ \midrule
\multicolumn{13}{c}{Table-as-an-image baseline} \\ \midrule
\multicolumn{1}{l|}{Gemini 1.5 Flash} &
  40.80 &
  44.22 &
  \multicolumn{1}{c|}{0.44} &
  22.30 &
  25.65 &
  \multicolumn{1}{c|}{0.34} &
  37.35 &
  41.01 &
  \multicolumn{1}{c|}{0.39} &
  33.20 &
  37.80 &
  \multicolumn{1}{c}{0.38}
   \\
\multicolumn{1}{l|}{GPT-4o} &
  \cellcolor[HTML]{FFCCC9}60.80 &
  \cellcolor[HTML]{FFCCC9}64.60 &
  \multicolumn{1}{c|}{\cellcolor[HTML]{FFCCC9}0.66} &
  \cellcolor[HTML]{FFCCC9}36.20 &
  \cellcolor[HTML]{FFCCC9}39.60 &
  \multicolumn{1}{c|}{\cellcolor[HTML]{FFCCC9}0.49} &
  \cellcolor[HTML]{FFCCC9}65.40 &
  \cellcolor[HTML]{FFCCC9}67.00 &
  \multicolumn{1}{c|}{\cellcolor[HTML]{FFCCC9}0.67} &
  \cellcolor[HTML]{FFCCC9}50.20 &
  \cellcolor[HTML]{FFCCC9}51.80 &
  \multicolumn{1}{c}{\cellcolor[HTML]{FFCCC9}0.50}
   \\
\multicolumn{1}{l|}{Qwen-VL-chat} &
  12.20 &
  14.04 &
  \multicolumn{1}{c|}{0.14} &
  3.60 &
  4.50 &
  \multicolumn{1}{c|}{0.07} &
  7.99 &
  9.38 &
  \multicolumn{1}{c|}{0.09} &
  10.40 &
  12.00 &
  \multicolumn{1}{c}{0.13}
   \\
\multicolumn{1}{l|}{CogAgent-VQA} &
  12.62 &
  14.89 &
  \multicolumn{1}{c|}{0.15} &
  5.59 &
  5.95 &
  \multicolumn{1}{c|}{0.10} &
  8.68 &
  11.28 &
  \multicolumn{1}{c|}{0.10} &
  8.40 &
  9.40 &
  \multicolumn{1}{c}{0.09}
   \\
\multicolumn{1}{l|}{Intern-VLM-4khd} &
  22.55 &
  26.67 &
  \multicolumn{1}{c|}{0.27} &
  11.35 &
  13.87 &
  \multicolumn{1}{c|}{0.20} &
  18.58 &
  22.22 &
  \multicolumn{1}{c|}{0.22} &
  15.40 &
  17.20 &
  \multicolumn{1}{c}{0.18}
   \\ \midrule
\multicolumn{13}{c}{Interleaved Image-text Baseline} \\ \midrule
\multicolumn{1}{l|}{Gemini 1.5 Flash} &
  47.98 &
  60.31 &
  \multicolumn{1}{c|}{0.53} &
  20.33 &
  33.33 &
  \multicolumn{1}{c|}{0.36} &
  44.21 &
  50.45 &
  \multicolumn{1}{c|}{0.47} &
  38.58 &
  \cellcolor[HTML]{FFCCC9}50.39 &
  \multicolumn{1}{c}{0.45}
   \\
\multicolumn{1}{l|}{GPT4o} &
  \cellcolor[HTML]{FFCCC9}69.65 &
  \cellcolor[HTML]{FFCCC9}72.47 &
  \multicolumn{1}{c|}{\cellcolor[HTML]{FFCCC9}0.72} &
  \cellcolor[HTML]{FFCCC9}44.44 &
  \cellcolor[HTML]{FFCCC9}49.27 &
  \multicolumn{1}{c|}{\cellcolor[HTML]{FFCCC9}0.57} &
  \cellcolor[HTML]{FFCCC9}68 &
  \cellcolor[HTML]{FFCCC9}69.6 &
  \multicolumn{1}{c|}{\cellcolor[HTML]{FFCCC9}0.69} &
  \cellcolor[HTML]{FFCCC9}46.40 &
  47.60 &
  \multicolumn{1}{c}{\cellcolor[HTML]{FFCCC9}0.49}
   \\
\multicolumn{1}{l|}{Qwen-VL} &
  1.04 &
  12.87 &
  \multicolumn{1}{c|}{0.02} &
  0.00 &
  6.65 &
  \multicolumn{1}{c|}{0.02} &
  0.19 &
  11.62 &
  \multicolumn{1}{c|}{0.01} &
  0.42 &
  10.29 &
  \multicolumn{1}{c}{0.02}
   \\
\multicolumn{1}{l|}{Idefics-Mantis} &
  5.44 &
  10.46 &
  \multicolumn{1}{c|}{0.07} &
  2.33 &
  2.62 &
  \multicolumn{1}{c|}{0.04} &
  3.46 &
  10.39 &
  \multicolumn{1}{c|}{0.06} &
  3.77 &
  8.49 &
  \multicolumn{1}{c}{0.05}
   \\
  
\end{tabular}%
}
\caption{Detailed results on sampled subset of WikiTableQuestions. EM - Exact Match, SSM - Substring Match, F1 - F1 score. EQ - Explicit Questions, AQ - Answer-Mention Questions, IQ - Implicit Questions, VQ - Visual Questions. Best performing models are highlighted in red.}
\label{tab:my-WTQ}
\end{table*}
% Please add the following required packages to your document preamble:
% \usepackage{graphicx}
% \usepackage[table,xcdraw]{xcolor}
% Beamer presentation requires \usepackage{colortbl} instead of \usepackage[table,xcdraw]{xcolor}
\begin{table*}[ht]
\small
\setlength{\tabcolsep}{3.5pt}
\resizebox{\textwidth}{!}{%
\begin{tabular}{@{}lllllllllllll@{}}
\toprule
\multicolumn{1}{l|}{Question-Type} &
  \multicolumn{3}{c|}{EQ} &
  \multicolumn{3}{c|}{IQ} &
  \multicolumn{3}{c|}{AQ} &
  \multicolumn{3}{c}{VQ}
   \\ 
   \midrule
\multicolumn{1}{l|}{Metrics} &
  \multicolumn{1}{c|}{EM} &
  \multicolumn{1}{c|}{SSM} &
  \multicolumn{1}{c|}{F1} &
  \multicolumn{1}{c|}{EM} &
  \multicolumn{1}{c|}{SSM} &
  \multicolumn{1}{c|}{F1} &
  \multicolumn{1}{c|}{EM} &
  \multicolumn{1}{c|}{SSM} &
  \multicolumn{1}{c|}{F1} &
  \multicolumn{1}{c|}{EM} &
  \multicolumn{1}{c|}{SSM} &
  \multicolumn{1}{c}{F1}
   \\ \midrule
\multicolumn{13}{c}{Partial Input Baseline} \\ \midrule
\multicolumn{1}{l|}{Gemini 1.5 Flash} &
  \cellcolor[HTML]{FFFFFF}36.43 &
  \cellcolor[HTML]{FFFFFF}39.14 &
  \multicolumn{1}{r|}{\cellcolor[HTML]{FFFFFF}0.37} &
  \cellcolor[HTML]{FFFFFF}26.71 &
  \cellcolor[HTML]{FFFFFF}28.71 &
  \multicolumn{1}{r|}{\cellcolor[HTML]{FFFFFF}0.38} &
  \cellcolor[HTML]{FFCCC9}{\color[HTML]{000000} 57.78} &
  \cellcolor[HTML]{FFCCC9}{\color[HTML]{000000} 62.22} &
  \multicolumn{1}{r|}{\cellcolor[HTML]{FFFFFF}0.21} &
  \cellcolor[HTML]{FFFFFF}24.60 &
  \cellcolor[HTML]{FFFFFF}28.00 &
  \multicolumn{1}{r}{\cellcolor[HTML]{FFFFFF}0.30}
   \\
\multicolumn{1}{l|}{LLAMA 3 70B} &
  \cellcolor[HTML]{FFCCC9}{\color[HTML]{000000} 38.25} &
  \cellcolor[HTML]{FFCCC9}{\color[HTML]{000000} 41.12} &
  \multicolumn{1}{r|}{\cellcolor[HTML]{FFCCC9}{\color[HTML]{000000} 0.38}} &
  \cellcolor[HTML]{FFCCC9}{\color[HTML]{000000} 27.75} &
  \cellcolor[HTML]{FFCCC9}{\color[HTML]{000000} 30.76} &
  \multicolumn{1}{r|}{\cellcolor[HTML]{FFCCC9}{\color[HTML]{000000} 0.41}} &
  \cellcolor[HTML]{FFFFFF}{\color[HTML]{000000} 56.83} &
  \cellcolor[HTML]{FFFFFF}{\color[HTML]{000000} 61.27} &
  \multicolumn{1}{r|}{\cellcolor[HTML]{FFCCC9}{\color[HTML]{000000} 0.57}} &
  \cellcolor[HTML]{FFCCC9}{\color[HTML]{000000} 27.80} &
  \cellcolor[HTML]{FFCCC9}{\color[HTML]{000000} 30.60} &
  \multicolumn{1}{r}{\cellcolor[HTML]{FFCCC9}{\color[HTML]{000000} 0.33}} \\
\multicolumn{1}{l|}{Mixtral 8x7B} &
  \cellcolor[HTML]{FFFFFF}20.86 &
  \cellcolor[HTML]{FFFFFF}23.43 &
  \multicolumn{1}{r|}{\cellcolor[HTML]{FFFFFF}0.22} &
  \cellcolor[HTML]{FFFFFF}13.86 &
  \cellcolor[HTML]{FFFFFF}17.71 &
  \multicolumn{1}{r|}{\cellcolor[HTML]{FFFFFF}0.24} &
  \cellcolor[HTML]{FFFFFF}24.13 &
  \cellcolor[HTML]{FFFFFF}28.89 &
  \multicolumn{1}{r|}{\cellcolor[HTML]{FFFFFF}0.25} &
  \cellcolor[HTML]{FFFFFF}15.00 &
  \cellcolor[HTML]{FFFFFF}19.20 &
  \multicolumn{1}{r}{\cellcolor[HTML]{FFFFFF}0.21}
   \\ \midrule
\multicolumn{13}{c}{Oracle-Entity Replaced Baseline} \\ \midrule
\multicolumn{1}{l|}{Gemini 1.5 Flash} &
  \cellcolor[HTML]{FFCCC9}79.00 &
  \cellcolor[HTML]{FFCCC9}82.29 &
  \multicolumn{1}{r|}{\cellcolor[HTML]{FFCCC9}0.73} &
  \cellcolor[HTML]{FFCCC9}80.57 &
  \cellcolor[HTML]{FFCCC9}81.86 &
  \multicolumn{1}{r|}{\cellcolor[HTML]{FFCCC9}0.90} &
  \cellcolor[HTML]{FFCCC9}73.02 &
  \cellcolor[HTML]{FFCCC9}77.46 &
  \multicolumn{1}{r|}{\cellcolor[HTML]{FFCCC9}0.72} &
  \multicolumn{1}{l}{\cellcolor[HTML]{FFFFFF}\textbf{-}} &
  \multicolumn{1}{l}{\cellcolor[HTML]{FFFFFF}\textbf{-}} &
  \multicolumn{1}{l}{\cellcolor[HTML]{FFFFFF}\textbf{-}}
   \\
\multicolumn{1}{l|}{LLAMA 3.00 70B} &
  \cellcolor[HTML]{FFFFFF}74.29 &
  \cellcolor[HTML]{FFFFFF}78.29 &
  \multicolumn{1}{r|}{\cellcolor[HTML]{FFFFFF}0.70} &
  \cellcolor[HTML]{FFFFFF}77.00 &
  \cellcolor[HTML]{FFFFFF}78.57 &
  \multicolumn{1}{r|}{\cellcolor[HTML]{FFFFFF}0.88} &
  \cellcolor[HTML]{FFFFFF}62.86 &
  \cellcolor[HTML]{FFFFFF}68.25 &
  \multicolumn{1}{r|}{\cellcolor[HTML]{FFFFFF}0.63} &
  \multicolumn{1}{l}{\cellcolor[HTML]{FFFFFF}\textbf{-}} &
  \multicolumn{1}{l}{\cellcolor[HTML]{FFFFFF}\textbf{-}} &
  \multicolumn{1}{l}{\cellcolor[HTML]{FFFFFF}\textbf{-}}
   \\
\multicolumn{1}{l|}{Mixtral 8x7B} &
  \cellcolor[HTML]{FFFFFF}54.71 &
  \cellcolor[HTML]{FFFFFF}59.29 &
  \multicolumn{1}{r|}{\cellcolor[HTML]{FFFFFF}0.55} &
  \cellcolor[HTML]{FFFFFF}60.71 &
  \cellcolor[HTML]{FFFFFF}65.29 &
  \multicolumn{1}{r|}{\cellcolor[HTML]{FFFFFF}0.75} &
  \cellcolor[HTML]{FFFFFF}27.94 &
  \cellcolor[HTML]{FFFFFF}33.97 &
  \multicolumn{1}{r|}{\cellcolor[HTML]{FFFFFF}0.28} &
  \multicolumn{1}{l}{\cellcolor[HTML]{FFFFFF}\textbf{-}} &
  \multicolumn{1}{l}{\cellcolor[HTML]{FFFFFF}\textbf{-}} &
  \multicolumn{1}{l}{\cellcolor[HTML]{FFFFFF}\textbf{-}}
   \\ \midrule
\multicolumn{13}{c}{\cellcolor[HTML]{FFFFFF}Image Captioning Baseline} \\ \midrule
\multicolumn{1}{l|}{Gemini 1.5 Flash} &
  \cellcolor[HTML]{FFFFFF}45.43 &
  \cellcolor[HTML]{FFFFFF}50.43 &
  \multicolumn{1}{r|}{\cellcolor[HTML]{FFFFFF}0.45} &
  \cellcolor[HTML]{FFFFFF}33.71 &
  \cellcolor[HTML]{FFFFFF}40.86 &
  \multicolumn{1}{r|}{\cellcolor[HTML]{FFFFFF}0.53} &
  \cellcolor[HTML]{FFFFFF}62.54 &
  \cellcolor[HTML]{FFFFFF}67.30 &
  \multicolumn{1}{r|}{\cellcolor[HTML]{FFFFFF}0.65} &
  \cellcolor[HTML]{FFFFFF}40.80 &
  \cellcolor[HTML]{FFFFFF}46.60 &
  \multicolumn{1}{r}{\cellcolor[HTML]{FFFFFF}0.45}
   \\ \midrule
\multicolumn{13}{c}{Table-as-an-image baseline} \\ \midrule
\multicolumn{1}{l|}{Gemini 1.5 flash} &
  \cellcolor[HTML]{FFFFFF}43.63 &
  \cellcolor[HTML]{FFFFFF}47.08 &
  \multicolumn{1}{r|}{\cellcolor[HTML]{FFFFFF}0.43} &
  \cellcolor[HTML]{FFFFFF}28.27 &
  \cellcolor[HTML]{FFFFFF}35.75 &
  \multicolumn{1}{r|}{\cellcolor[HTML]{FFFFFF}0.47} &
  \cellcolor[HTML]{FFFFFF}46.03 &
  \cellcolor[HTML]{FFFFFF}52.38 &
  \multicolumn{1}{r|}{\cellcolor[HTML]{FFFFFF}0.49} &
  \cellcolor[HTML]{FFFFFF}32.51 &
  \cellcolor[HTML]{FFFFFF}35.25 &
  \multicolumn{1}{r}{\cellcolor[HTML]{FFFFFF}0.35}
   \\
\multicolumn{1}{l|}{GPT-4o} &
  \cellcolor[HTML]{FFCCC9}51.60 &
  \cellcolor[HTML]{FFCCC9}55.00 &
  \multicolumn{1}{r|}{\cellcolor[HTML]{FFCCC9}0.48} &
  \cellcolor[HTML]{FFCCC9}39.00 &
  \cellcolor[HTML]{FFCCC9}43.20 &
  \multicolumn{1}{r|}{\cellcolor[HTML]{FFCCC9}0.59} &
  \cellcolor[HTML]{FFCCC9}58.73 &
  \cellcolor[HTML]{FFCCC9}62.22 &
  \multicolumn{1}{r|}{\cellcolor[HTML]{FFCCC9}0.60} &
  \cellcolor[HTML]{FFCCC9}47.80 &
  \cellcolor[HTML]{FFCCC9}54.40 &
  \multicolumn{1}{r}{\cellcolor[HTML]{FFCCC9}0.52}
   \\
\multicolumn{1}{l|}{Qwen-VL-chat} &
  \cellcolor[HTML]{FFFFFF}6.29 &
  \cellcolor[HTML]{FFFFFF}9.59 &
  \multicolumn{1}{r|}{\cellcolor[HTML]{FFFFFF}0.10} &
  \cellcolor[HTML]{FFFFFF}6.20 &
  \cellcolor[HTML]{FFFFFF}7.14 &
  \multicolumn{1}{r|}{\cellcolor[HTML]{FFFFFF}0.12} &
  \cellcolor[HTML]{FFFFFF}17.14 &
  \cellcolor[HTML]{FFFFFF}35.24 &
  \multicolumn{1}{r|}{\cellcolor[HTML]{FFFFFF}0.19} &
  \cellcolor[HTML]{FFFFFF}4.60 &
  \cellcolor[HTML]{FFFFFF}8.40 &
  \multicolumn{1}{r}{\cellcolor[HTML]{FFFFFF}0.08}
   \\
\multicolumn{1}{l|}{CogAgent-VQA} &
  \cellcolor[HTML]{FFFFFF}10.07 &
  \cellcolor[HTML]{FFFFFF}13.08 &
  \multicolumn{1}{r|}{\cellcolor[HTML]{FFFFFF}0.14} &
  \cellcolor[HTML]{FFFFFF}10.27 &
  \cellcolor[HTML]{FFFFFF}11.53 &
  \multicolumn{1}{r|}{\cellcolor[HTML]{FFFFFF}0.16} &
  \cellcolor[HTML]{FFFFFF}6.98 &
  \cellcolor[HTML]{FFFFFF}19.37 &
  \multicolumn{1}{r|}{\cellcolor[HTML]{FFFFFF}0.07} &
  \cellcolor[HTML]{FFFFFF}5.80 &
  \cellcolor[HTML]{FFFFFF}8.80 &
  \multicolumn{1}{r}{\cellcolor[HTML]{FFFFFF}0.09} 
   \\
\multicolumn{1}{l|}{Intern-VLM-4khd} &
  \cellcolor[HTML]{FFFFFF}24.00 &
  \cellcolor[HTML]{FFFFFF}28.71 &
  \multicolumn{1}{r|}{\cellcolor[HTML]{FFFFFF}0.26} &
  \cellcolor[HTML]{FFFFFF}15.14 &
  \cellcolor[HTML]{FFFFFF}18.00 &
  \multicolumn{1}{r|}{\cellcolor[HTML]{FFFFFF}0.27} &
  \cellcolor[HTML]{FFFFFF}23.49 &
  \cellcolor[HTML]{FFFFFF}29.84 &
  \multicolumn{1}{r|}{\cellcolor[HTML]{FFFFFF}0.26} &
  \cellcolor[HTML]{FFFFFF}5.80 &
  \cellcolor[HTML]{FFFFFF}9.60 &
  \multicolumn{1}{r}{\cellcolor[HTML]{FFFFFF}0.11}
   \\ \midrule
\multicolumn{13}{c}{Interleaved Image-text Baseline} \\ 
\midrule
\multicolumn{1}{l|}{Gemini 1.5 - Flash} &
  \cellcolor[HTML]{FFFFFF}56.95 &
  \cellcolor[HTML]{FFFFFF}53.22 &
  \multicolumn{1}{r|}{\cellcolor[HTML]{FFFFFF}0.54} &
  \cellcolor[HTML]{FFFFFF}32.59 &
  \cellcolor[HTML]{FFFFFF}40.18 &
  \multicolumn{1}{r|}{\cellcolor[HTML]{FFFFFF}0.49} &
  \cellcolor[HTML]{FFFFFF}53.23 &
  \cellcolor[HTML]{FFCCC9}62.90 &
  \multicolumn{1}{r|}{\cellcolor[HTML]{FFFFFF}0.54} &
  \cellcolor[HTML]{FFFFFF}43.61 &
  \cellcolor[HTML]{FFFFFF}48.02 &
  \multicolumn{1}{r}{\cellcolor[HTML]{FFFFFF}0.49}
   \\
\multicolumn{1}{l|}{GPT4o} &
  \cellcolor[HTML]{FFCCC9}63.00 &
  \cellcolor[HTML]{FFCCC9}66.50 &
  \multicolumn{1}{r|}{\cellcolor[HTML]{FFCCC9}0.62} &
  \cellcolor[HTML]{FFCCC9}39.96 &
  \cellcolor[HTML]{FFCCC9}48.93 &
  \multicolumn{1}{r|}{\cellcolor[HTML]{FFCCC9}0.61} &
  \cellcolor[HTML]{FFCCC9}55.24 &
  \cellcolor[HTML]{FFFFFF}57.78 &
  \multicolumn{1}{r|}{\cellcolor[HTML]{FFCCC9}0.56} &
  \cellcolor[HTML]{FFCCC9}48.60 &
  \cellcolor[HTML]{FFCCC9}54.00 &
  \multicolumn{1}{r}{\cellcolor[HTML]{FFCCC9}0.49}
   \\
\multicolumn{1}{l|}{Qwen-VL} &
  \cellcolor[HTML]{FFFFFF}2.63 &
  \cellcolor[HTML]{FFFFFF}9.60 &
  \multicolumn{1}{r|}{\cellcolor[HTML]{FFFFFF}0.05} &
  \cellcolor[HTML]{FFFFFF}1.82 &
  \cellcolor[HTML]{FFFFFF}5.38 &
  \multicolumn{1}{r|}{\cellcolor[HTML]{FFFFFF}0.04} &
  \cellcolor[HTML]{FFFFFF}5.08 &
  \cellcolor[HTML]{FFFFFF}12.88 &
  \multicolumn{1}{r|}{\cellcolor[HTML]{FFFFFF}0.06} &
  \cellcolor[HTML]{FFFFFF}1.14 &
  \cellcolor[HTML]{FFFFFF}7.09 &
  \multicolumn{1}{r}{\cellcolor[HTML]{FFFFFF}0.03}
   \\
\multicolumn{1}{l|}{Idefics-Mantis} &
  \cellcolor[HTML]{FFFFFF}1.33 &
  \cellcolor[HTML]{FFFFFF}2.88 &
  \multicolumn{1}{r|}{\cellcolor[HTML]{FFFFFF}0.03} &
  \cellcolor[HTML]{FFFFFF}4.85 &
  \cellcolor[HTML]{FFFFFF}5.70 &
  \multicolumn{1}{r|}{\cellcolor[HTML]{FFFFFF}0.08} &
  \cellcolor[HTML]{FFFFFF}0.00 &
  \cellcolor[HTML]{FFFFFF}9.09 &
  \multicolumn{1}{r|}{\cellcolor[HTML]{FFFFFF}0.04} &
  \cellcolor[HTML]{FFFFFF}1.08 &
  \cellcolor[HTML]{FFFFFF}3.61 &
  \multicolumn{1}{r}{\cellcolor[HTML]{FFFFFF}0.03}
  
\end{tabular}%
}
\caption{Detailed results on sampled subset of WikiSQL. EM - Exact Match, SSM - Substring Match, F1 - F1 score. EQ - Explicit Questions, AQ - Answer-Mention Questions, IQ - Implicit Questions, VQ - Visual Questions. Best performing models are highlighted in red.}
\label{tab:WSL}
\end{table*}
% Please add the following required packages to your document preamble:
% \usepackage{booktabs}
% \usepackage{graphicx}
% \usepackage[table,xcdraw]{xcolor}
% Beamer presentation requires \usepackage{colortbl} instead of \usepackage[table,xcdraw]{xcolor}
\begin{table*}[ht]
\small
\setlength{\tabcolsep}{2.5pt}
\resizebox{\textwidth}{!}{%
\begin{tabular}{@{}lllllllllllllllll@{}}
\toprule
\multicolumn{1}{l|}{Models} &
  \multicolumn{4}{c|}{EQ} &
  \multicolumn{4}{c|}{AQ} &
  \multicolumn{4}{c|}{IQ} &
  \multicolumn{4}{c}{VQ} \\ \midrule
\multicolumn{1}{l|}{Metrics} &
  \multicolumn{1}{l|}{BLEU} &
  \multicolumn{1}{l|}{R-1} &
  \multicolumn{1}{l|}{R-2} &
  \multicolumn{1}{l|}{R-L} &
  \multicolumn{1}{l|}{BLEU} &
  \multicolumn{1}{l|}{R-1} &
  \multicolumn{1}{l|}{R-2} &
  \multicolumn{1}{l|}{R-L} &
  \multicolumn{1}{l|}{BLEU} &
  \multicolumn{1}{l|}{R-1} &
  \multicolumn{1}{l|}{R-2} &
  \multicolumn{1}{l|}{R-L} &
  \multicolumn{1}{l|}{BLEU} &
  \multicolumn{1}{l|}{R-1} &
  \multicolumn{1}{l|}{R-2} &
  R-L \\ \midrule
\multicolumn{17}{c}{Oracle-Entity Replaced Baseline} \\ \midrule
Gemini 1.5 Flash &
  \cellcolor[HTML]{FFCCC9}29.93 &
  0.62 &
  0.42 &
  0.53 &
  \cellcolor[HTML]{FFCCC9}23.29 &
  \cellcolor[HTML]{FFCCC9}0.62 &
  \cellcolor[HTML]{FFCCC9}0.39 &
  \cellcolor[HTML]{FFCCC9}0.50 &
  \cellcolor[HTML]{FFCCC9}16.37 &
  \cellcolor[HTML]{FFCCC9}0.50 &
  \cellcolor[HTML]{FFCCC9}0.28 &
  \cellcolor[HTML]{FFCCC9}0.41 &
  - &
  - &
  - &
  - \\
LLAMA 3 70B &
  29.13 &
  \cellcolor[HTML]{FFCCC9}0.66 &
  \cellcolor[HTML]{FFCCC9}0.45 &
  \cellcolor[HTML]{FFCCC9}0.54 &
  19.23 &
  0.58 &
  0.36 &
  0.46 &
  15.10 &
  0.50 &
  0.28 &
  0.41 &
  - &
  - &
  - &
  - \\
Mixtral 8x7B &
  7.97 &
  0.49 &
  0.31 &
  0.42 &
  11.16 &
  0.51 &
  0.31 &
  0.41 &
  7.54 &
  0.41 &
  0.22 &
  0.33 &
  - &
  - &
  - &
  - \\ \midrule
\multicolumn{17}{c}{Partial Input Baseline} \\ \midrule
Gemini 1.5 Flash &
  16.24 &
  0.52 &
  0.33 &
  0.44 &
  20.17 &
  0.54 &
  0.31 &
  0.44 &
  \cellcolor[HTML]{FFCCC9}20.39 &
  0.53 &
  0.31 &
  0.44 &
  22.89 &
  0.55 &
  0.34 &
  0.47 \\
LLAMA 3 70B &
  \cellcolor[HTML]{FFCCC9}31.49 &
  \cellcolor[HTML]{FFCCC9}0.63 &
  \cellcolor[HTML]{FFCCC9}0.42 &
  \cellcolor[HTML]{FFCCC9}0.53 &
  \cellcolor[HTML]{FFCCC9}21.88 &
  \cellcolor[HTML]{FFCCC9}0.56 &
  \cellcolor[HTML]{FFCCC9}0.33 &
  \cellcolor[HTML]{FFCCC9}0.46 &
  19.69 &
  \cellcolor[HTML]{FFCCC9}0.54 &
  \cellcolor[HTML]{FFCCC9}0.31 &
  \cellcolor[HTML]{FFCCC9}0.45 &
  \cellcolor[HTML]{FFCCC9}24.42 &
  \cellcolor[HTML]{FFCCC9}0.56 &
  \cellcolor[HTML]{FFCCC9}0.35 &
  \cellcolor[HTML]{FFCCC9}0.48 \\
Mixtral 8x7B &
  18.39 &
  0.54 &
  0.34 &
  0.44 &
  12.97 &
  0.48 &
  0.26 &
  0.39 &
  11.73 &
  0.47 &
  0.26 &
  0.38 &
  14.11 &
  0.48 &
  0.27 &
  0.39 \\ \midrule
\multicolumn{17}{c}{Image-Captioning Baseline} \\ \midrule
Gemini 1.5 flash &
  4.31 &
  0.57 &
  0.37 &
  0.48 &
  7.43 &
  0.56 &
  0.34 &
  0.46 &
  8.19 &
  0.50 &
  0.28 &
  0.42 &
  7.52 &
  0.51 &
  0.31 &
  0.43 \\ \midrule
\multicolumn{17}{c}{Table-as-an-Image} \\ \midrule
Gemini 1.5 flash &
  \cellcolor[HTML]{FFCCC9}29.93 &
  0.62 &
  0.42 &
  0.53 &
  18.73 &
  0.54 &
  0.32 &
  0.44 &
  7.09 &
  0.51 &
  0.30 &
  0.43 &
  13.46 &
  0.56 &
  0.34 &
  0.47 \\
GPT-4o &
  29.13 &
  \cellcolor[HTML]{FFCCC9}0.66 &
  \cellcolor[HTML]{FFCCC9}0.45 &
  \cellcolor[HTML]{FFCCC9}0.54 &
  \cellcolor[HTML]{FFCCC9}19.74 &
  \cellcolor[HTML]{FFCCC9}0.58 &
  \cellcolor[HTML]{FFCCC9}0.35 &
  \cellcolor[HTML]{FFCCC9}0.47 &
  \cellcolor[HTML]{FFCCC9}19.08 &
  \cellcolor[HTML]{FFCCC9}0.55 &
  \cellcolor[HTML]{FFCCC9}0.33 &
  \cellcolor[HTML]{FFCCC9}0.45 &
  \cellcolor[HTML]{FFCCC9}21.70 &
  \cellcolor[HTML]{FFCCC9}0.59 &
  \cellcolor[HTML]{FFCCC9}0.37 &
  \cellcolor[HTML]{FFCCC9}0.49 \\
Qwen-VL-chat &
  7.97 &
  0.49 &
  0.31 &
  0.42 &
  5.21 &
  0.40 &
  0.20 &
  0.33 &
  3.85 &
  0.38 &
  0.19 &
  0.31 &
  6.53 &
  0.43 &
  0.23 &
  0.36 \\
CogAgent-VQA &
  8.14 &
  0.46 &
  0.27 &
  0.37 &
  5.43 &
  0.36 &
  0.18 &
  0.29 &
  1.72 &
  0.19 &
  0.07 &
  0.15 &
  1.22 &
  0.15 &
  0.05 &
  0.12 \\
Intern-VLM-4khd &
  16.24 &
  0.52 &
  0.33 &
  0.44 &
  9.78 &
  0.44 &
  0.24 &
  0.36 &
  9.39 &
  0.40 &
  0.21 &
  0.32 &
  11.10 &
  0.41 &
  0.23 &
  0.35 \\ \midrule
\multicolumn{17}{c}{Interleaved Text-Image Baseline} \\ \midrule
Gemini 1.5 Flash &
  27.13 &
  0.57 &
  0.36 &
  0.52 &
  19.24 &
  0.48 &
  0.25 &
  0.43 &
  21.88 &
  0.48 &
  0.27 &
  0.42 &
  23.51 &
  0.56 &
  0.33 &
  \cellcolor[HTML]{FFCCC9}0.51 \\
GPT4o &
  \cellcolor[HTML]{FFCCC9}32.41 &
  \cellcolor[HTML]{FFCCC9}0.67 &
  \cellcolor[HTML]{FFCCC9}0.47 &
  \cellcolor[HTML]{FFCCC9}0.56 &
  \cellcolor[HTML]{FFCCC9}24.56 &
  \cellcolor[HTML]{FFCCC9}0.62 &
  \cellcolor[HTML]{FFCCC9}0.39 &
  \cellcolor[HTML]{FFCCC9}0.51 &
  \cellcolor[HTML]{FFCCC9}22.08 &
  \cellcolor[HTML]{FFCCC9}0.56 &
  \cellcolor[HTML]{FFCCC9}0.33 &
  \cellcolor[HTML]{FFCCC9}0.46 &
  \cellcolor[HTML]{FFCCC9}26.74 &
  \cellcolor[HTML]{FFCCC9}0.63 &
  \cellcolor[HTML]{FFCCC9}0.41 &
  0.49 \\
Qwen-VL &
  3.81 &
  0.19 &
  0.07 &
  0.16 &
  3.45 &
  0.20 &
  0.06 &
  0.17 &
  4.20 &
  0.17 &
  0.05 &
  0.05 &
  1.48 &
  0.11 &
  0.03 &
  0.10 \\
Idefics-Mantis &
  7.71 &
  0.38 &
  0.21 &
  0.35 &
  7.67 &
  0.22 &
  0.12 &
  0.20 &
  9.36 &
  0.34 &
  0.17 &
  0.31 &
  3.49 &
  0.33 &
  0.17 &
  0.31
\end{tabular}%
}
\caption{Detailed results on sampled subset of FeTaQA. BLEU, ROUGE-1,2,L are reported in the table. EQ - Explicit Questions, AQ - Answer-Mention Questions, IQ - Implicit Questions, VQ - Visual Questions. Best performing models are highlighted in red.}
\label{tab:table-fetaqa}
\end{table*}

\section{Visual Question Error Analysis}
Upon performing a further fine grained analysis of the incorrect marked samples we identify 5 types of errors in the visual questions (breakdown of the ~15\% questions) and suggest some ways of rectifying them in future:

\paragraph{Non-repairable (0.6\%)} – These are visual questions based on portraits/paintings where no unique visual attributes are present in the table. We can identify them by prompting a VLM to check for visual attributes present in multiple images and discard them if found.
\paragraph{Hallucinated attributes (6.6\%)} – These questions have partially or completely incorrect visual attributes for the entity. We can use a VLM to check if the attributes are present in the entity’s image and discard them if they aren’t.
\paragraph{Hard to identify attributes (1.8\%)} – These questions rely on visual attributes that are hard to spot in the table image and require a zoomed-in, high-resolution view. While technically correct, they aren't relevant for table question answering. A VLM can help identify and prune these by checking if the attributes are easily noticeable in the key entity's image.
\paragraph{Non-unique attributes (6.4\%)} – These questions involve non-unique visual attributes for the entity, but unlike non-repairable questions, a unique set of attributes is possible. We can identify them by using a VLM to check if the attribute appears in multiple images and discard those that do.
\paragraph{No visual attribute (0.6\%)} – These questions refer directly to the entity name, as in a logo or poster (e.g., "Star Wars" on a Star Wars poster). They can be filtered by checking if the question's tokens completely overlap with the entity name.

\section{Prompt Samples}
\begin{figure*}[p]
    \centering
    \includegraphics[height=4.1in]{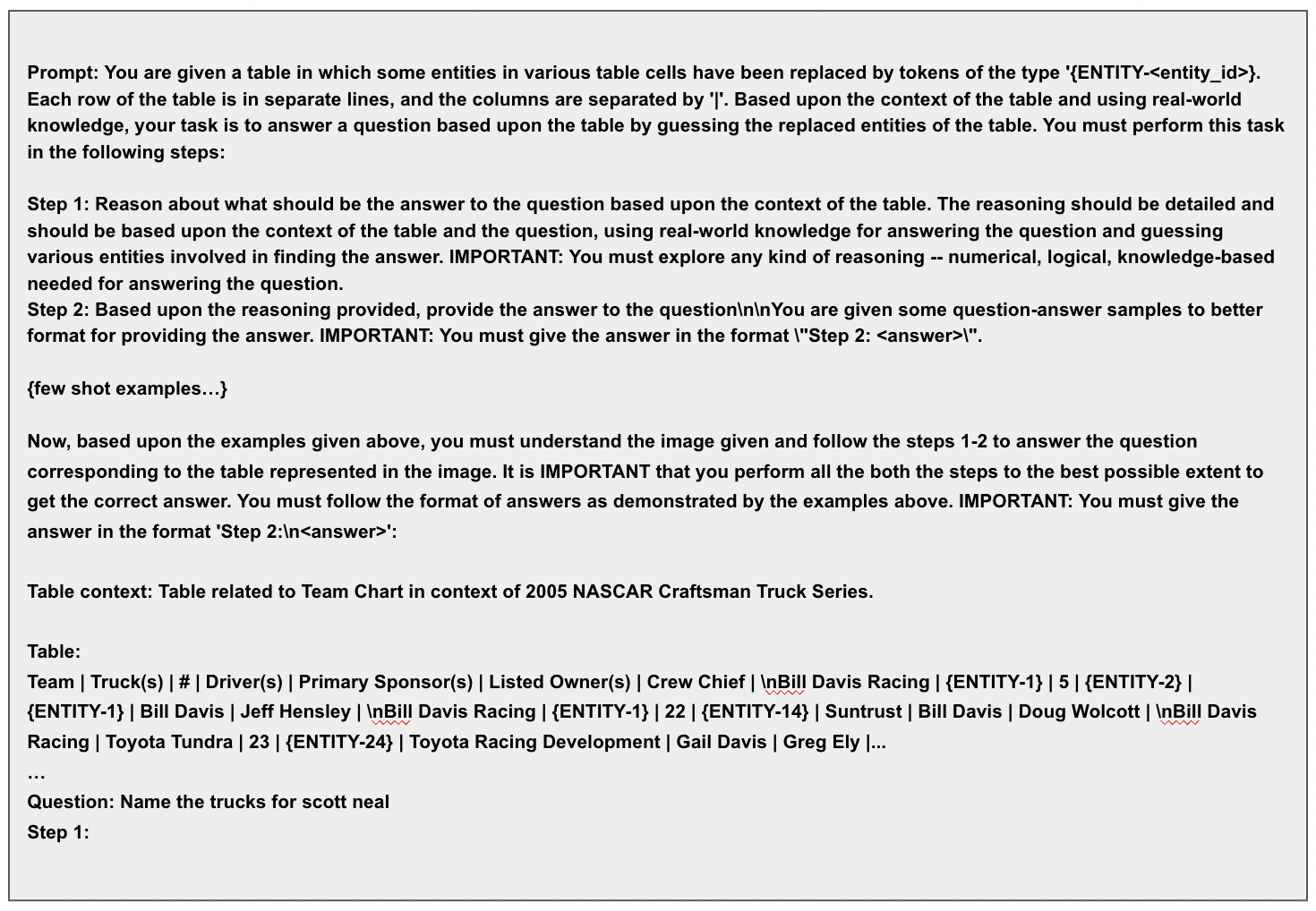}
    \caption{Prompt used for Partial Input Baseline}
    \label{fig:PI_prompt}
\end{figure*}

\begin{figure*}[ht]
    \centering
    \includegraphics[height=4.1in]{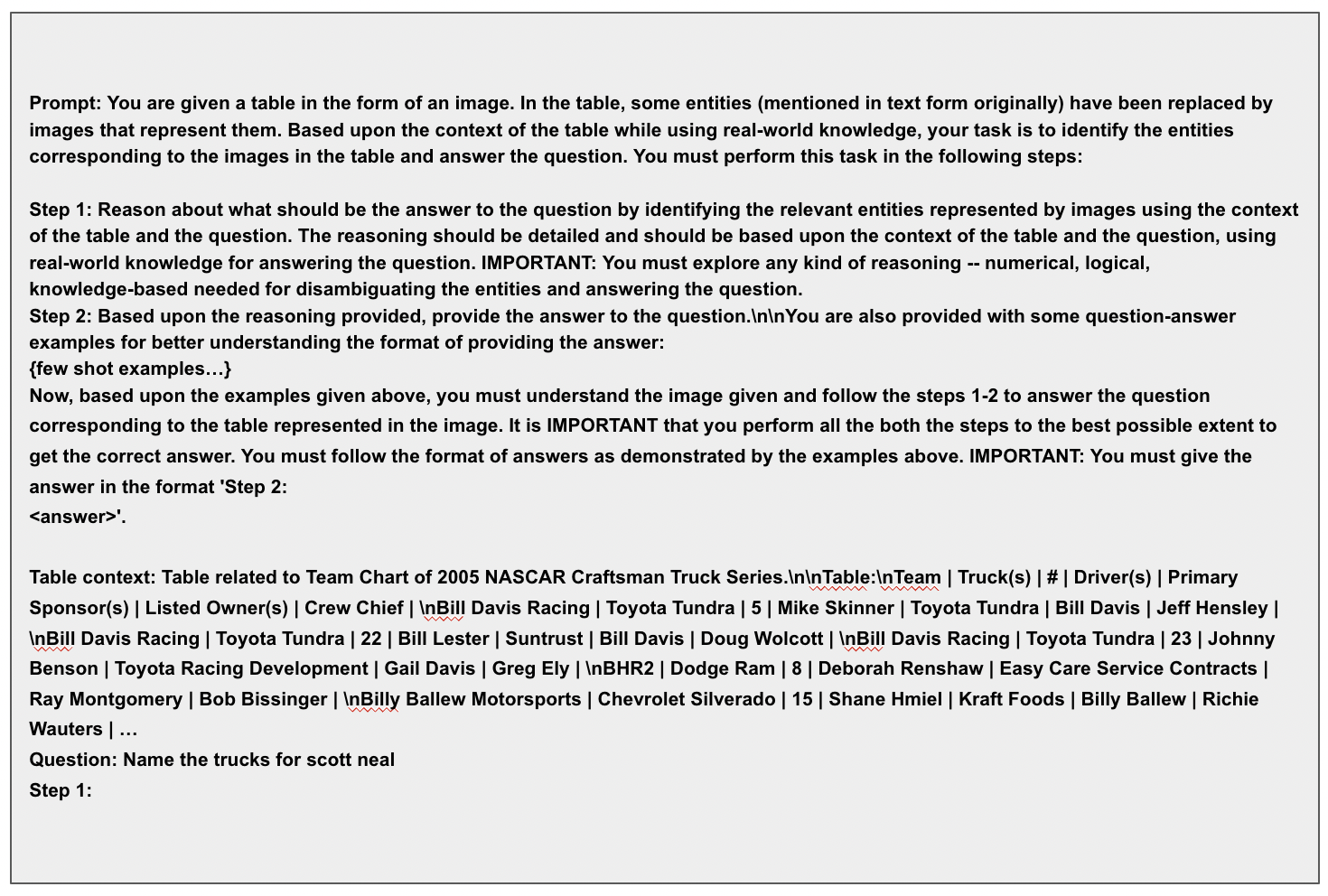}
    \caption{Prompt used for Oracle entity replacement Baseline}
    \label{fig:Oracle_prompt}
\end{figure*}

\begin{figure*}[ht]
    \centering
    \includegraphics[height=4.3in]{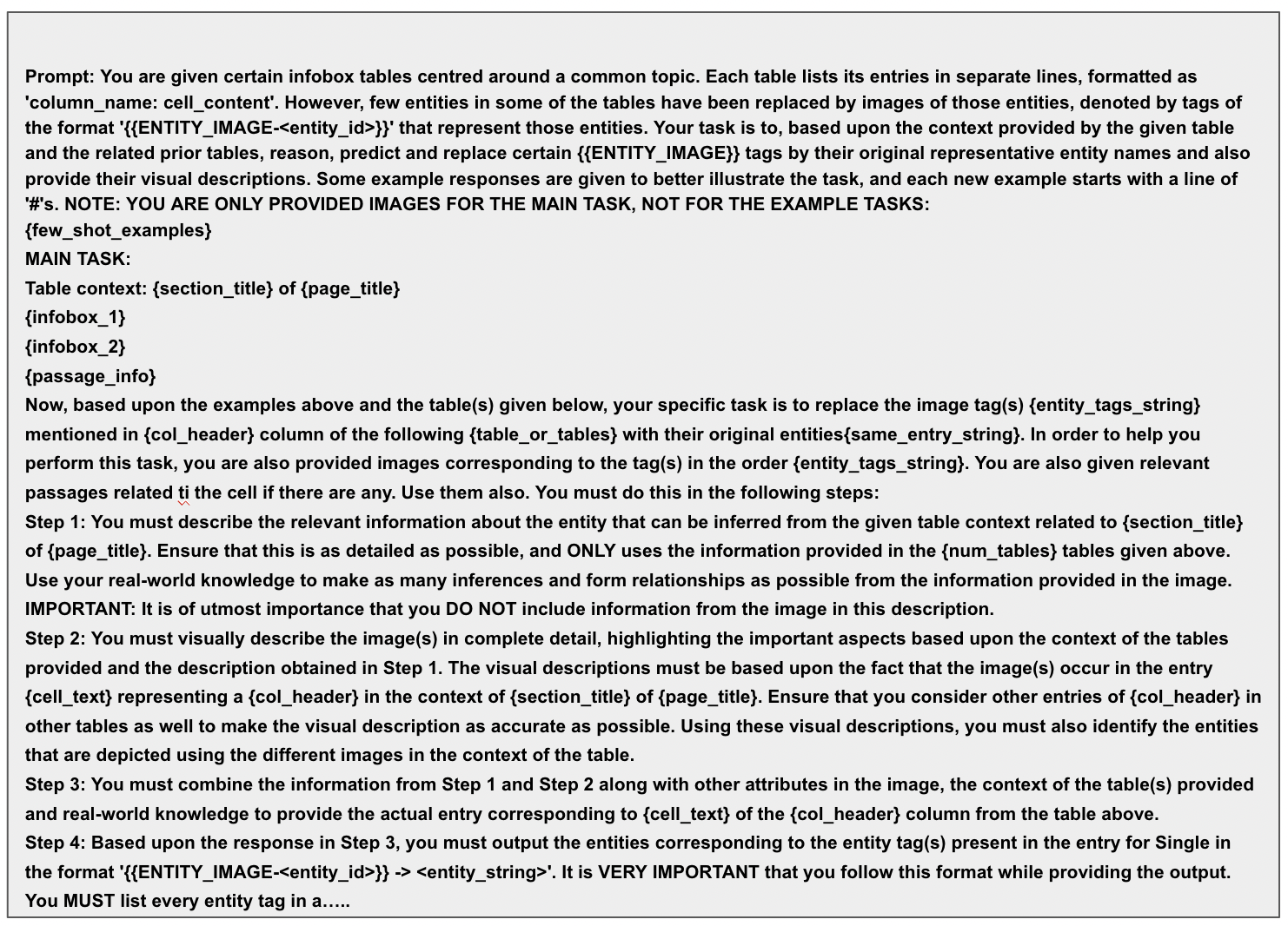}
    \caption{Prompt used for Image captioning Baseline}
    \label{fig:Image_captioning}
\end{figure*}

\begin{figure*}[ht]
    \centering
    \includegraphics[height=3.9in]{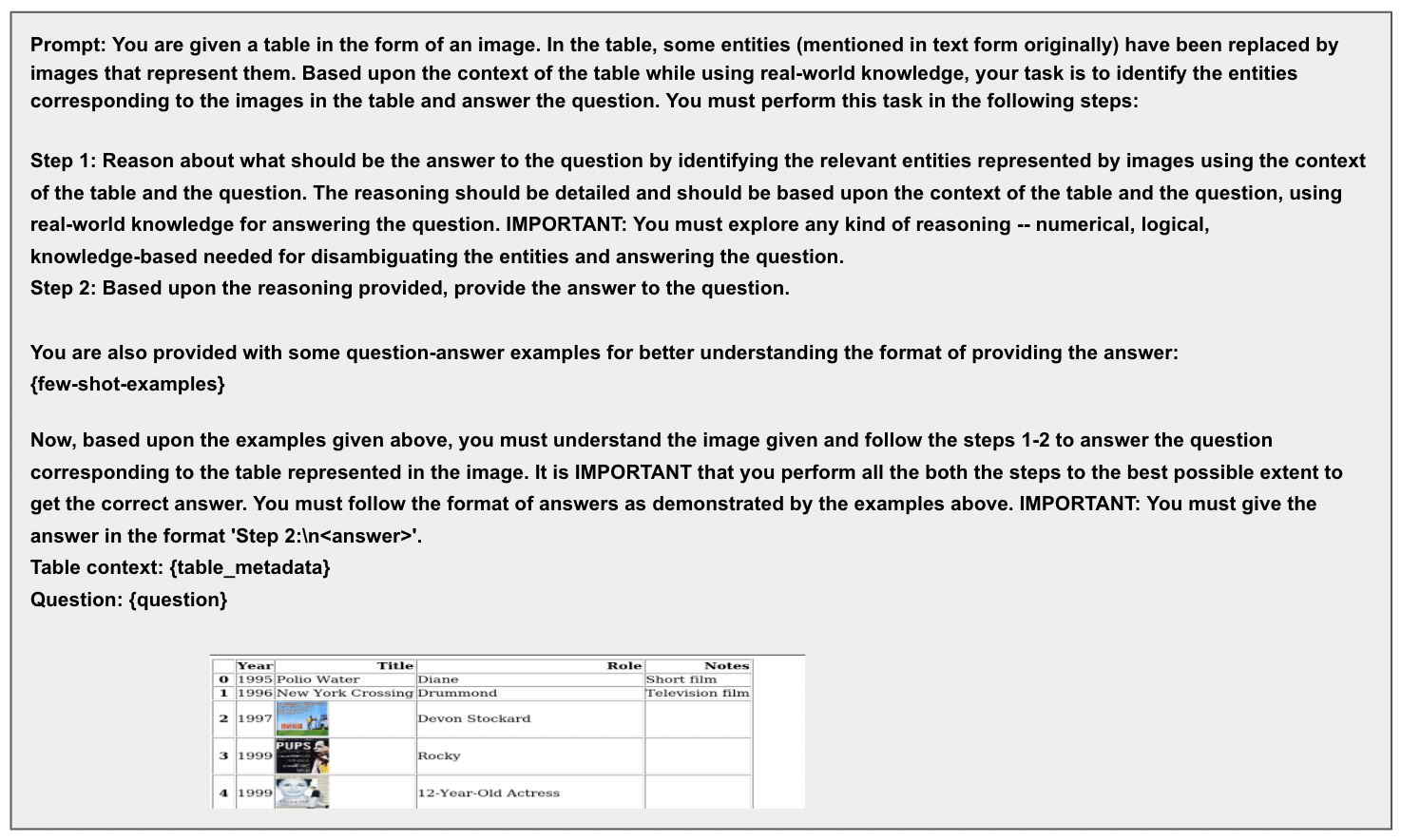}
    \caption{Prompt used for Table as an image baseline}
    \label{fig:/table_img}
\end{figure*}

\begin{figure*}[ht]
    \centering
    \includegraphics[height=4.1in]{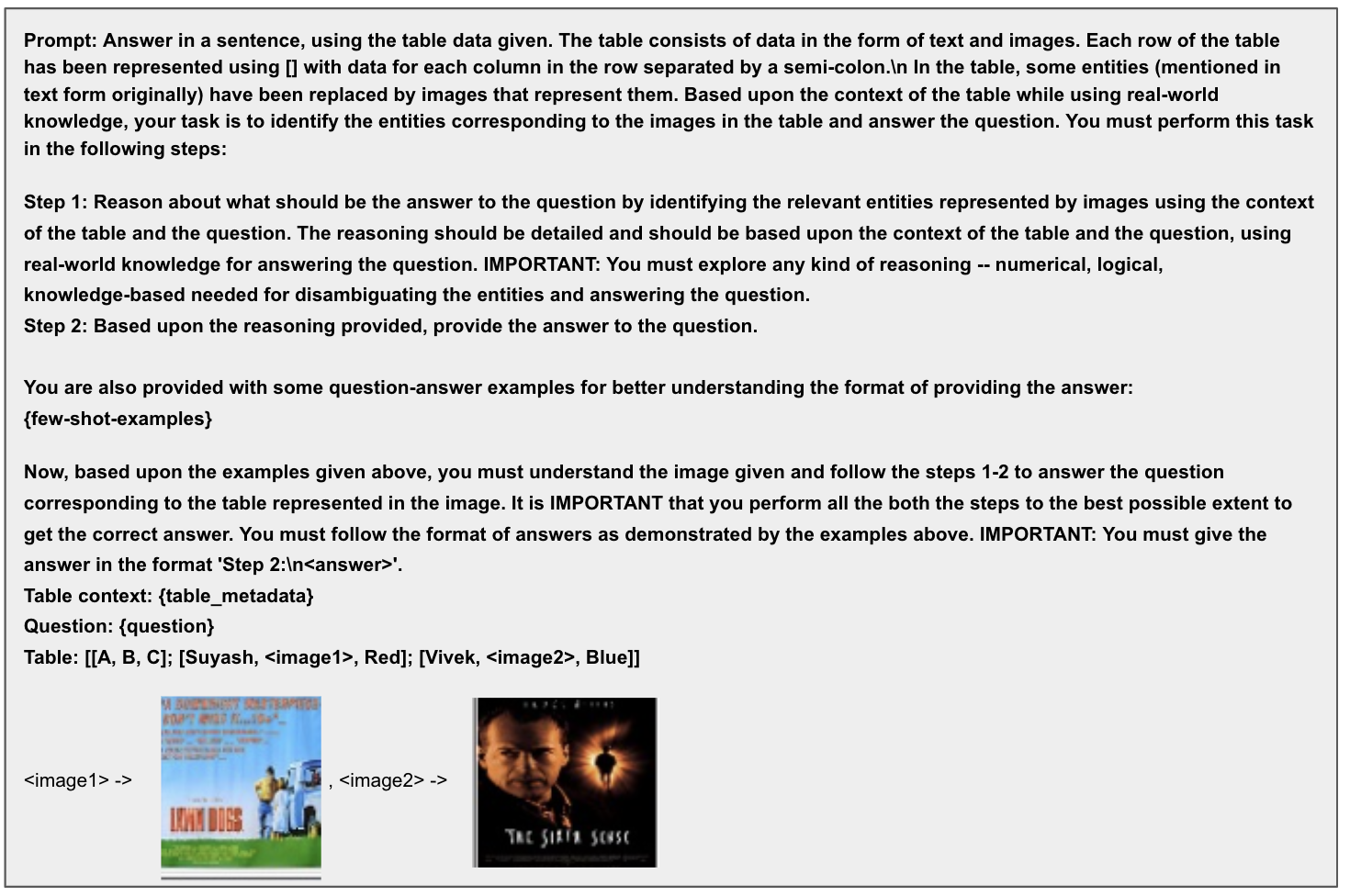}
    \caption{Prompt used for Interleaved Text-Image baseline}
    \label{fig:interleave}
\end{figure*}

We provide a detailed sample for each baseline strategy to illustrate our approach: Partial Input (Fig.~\ref{fig:PI_prompt}), Oracle-Based Entity (Fig.~\ref{fig:Oracle_prompt}), Image-Caption Fig.(~\ref{fig:Image_captioning}), Table-Image (Fig.~\ref{fig:/table_img}), and Interleaved Text-Image (Fig.~\ref{fig:interleave}). These examples demonstrate the varying degrees of information and context provided to the models, highlighting the differences in their ability to process and respond to diverse types of input. Through these samples, we aim to showcase the challenges and nuances involved in each strategy, offering insights into the models' performance.

% \section{Where Do Models Fail?}

% A significant observation from our experiments is the varied performance of models across different baseline settings and the reasons underlying these disparities. Gemini 1.5-flash demonstrates competitive performance across all data sources and methodologies. We identify four main issues: first, entity disambiguation, where the model inaccurately identifies entities from images, leading to errors. For instance, in Figure ~\autoref{fig:subfigures}(a), the model misidentifies the logo of California PA as that of the University of Lafayette. Secondly, the identification of visual attributes presents challenges for multimodal models, particularly in recognizing crucial visual elements within an image required to answer associated questions. For example, in ~\autoref{fig:subfigures}(c), the model's failure to correctly identify a flag based on its colors results in inaccurate responses. Thirdly, challenges arise in handling excessive content, leading to instances of incomplete or incorrect retrieval by the model. ~\autoref{fig:subfigures}(b) illustrates such a scenario where the model's incomplete understanding of data leads to erroneous conclusions. Lastly, despite correctly identifying entities at times, the model occasionally reaches incorrect conclusions, resulting in erroneous answers, as depicted in ~\autoref{fig:subfigures}(d).

\end{document}